\documentclass[10pt,twocolumn,letterpaper]{article}

\usepackage[pagenumbers]{cvpr} 

\usepackage{graphicx}
\usepackage{amsmath}
\usepackage{amssymb}
\usepackage{booktabs}

\usepackage{bbm}
\usepackage[ruled,linesnumbered]{algorithm2e}
\usepackage{setspace}

\usepackage{tabularx}
\usepackage{booktabs}
\usepackage{multirow}
\usepackage{diagbox}

\makeatletter
\@namedef{ver@everyshi.sty}{}
\makeatother
\usepackage{tikz}
\usetikzlibrary{spy}

\tikzset{
    cross/.pic = {
    \draw (-#1,0) -- (#1,0);
    \draw (0,-#1) -- (0,#1);
    }
}

\usepackage{xcolor}
\usepackage{color}
\usepackage{colortbl}

\definecolor{mygray}{gray}{.9}

\usepackage[accsupp]{axessibility}

\usepackage[american]{babel}

\makeatletter
\newcommand{\rmnum}[1]{\romannumeral #1}
\newcommand{\Rmnum}[1]{\expandafter\@slowromancap\romannumeral #1@}
\makeatother

\usepackage[pagebackref,breaklinks,colorlinks]{hyperref}

\usepackage[capitalize]{cleveref}
\crefname{section}{Sec.}{Secs.}
\Crefname{section}{Section}{Sections}
\Crefname{table}{Table}{Tables}
\crefname{table}{Tab.}{Tabs.}

\def\cvprPaperID{2709} 
\def\confName{CVPR}
\def\confYear{2022}

\begin{document}

\title{BokehMe: When Neural Rendering Meets Classical Rendering}

\author{Juewen Peng$^1$, Zhiguo Cao$^1$, Xianrui Luo$^1$, Hao Lu$^1$, Ke Xian$^1$\footnotemark[1]~, and Jianming Zhang$^2$ \\
$^1$Key Laboratory of Image Processing and Intelligent Control, Ministry of Education,\\
School of Artificial Intelligence and Automation, Huazhong University of Science and Technology\\
$^2$Adobe Research \\
{\tt\small \{juewenpeng, zgcao, xianruiluo, hlu, kexian\}@hust.edu.cn, jianmzha@adobe.com} \\
{\small{\url{https://github.com/JuewenPeng/BokehMe}}}
}

\twocolumn[{%
\renewcommand\twocolumn[1][]{#1}%
\maketitle

\begin{center}
\vspace{-15pt}
\setlength{\abovecaptionskip}{3pt}
\setlength{\belowcaptionskip}{5pt}
\captionsetup{type= float type}
\small
\centering
\setlength{\tabcolsep}{1pt}
\renewcommand{\arraystretch}{0.7}
\begin{tabular}{*{5}{c}}
    \includegraphics[trim={0 35pt 0 0},clip,height=2.7cm]{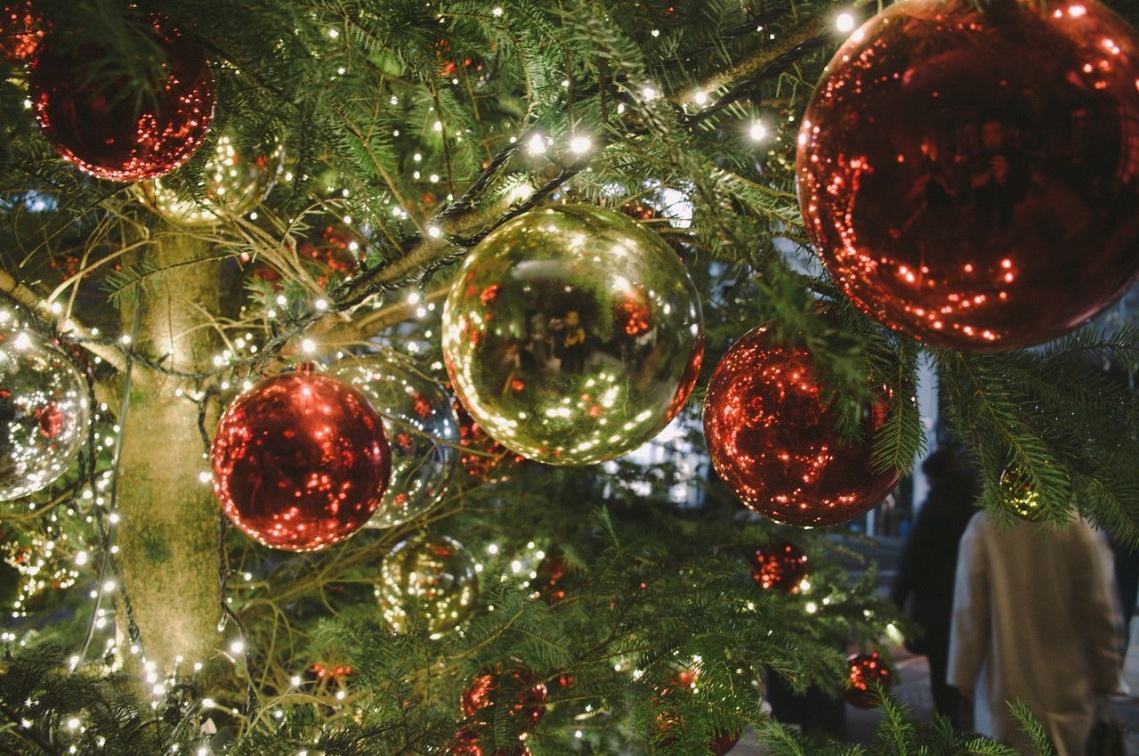} &
    \includegraphics[trim={0 35pt 0 245pt},clip,height=2.7cm]{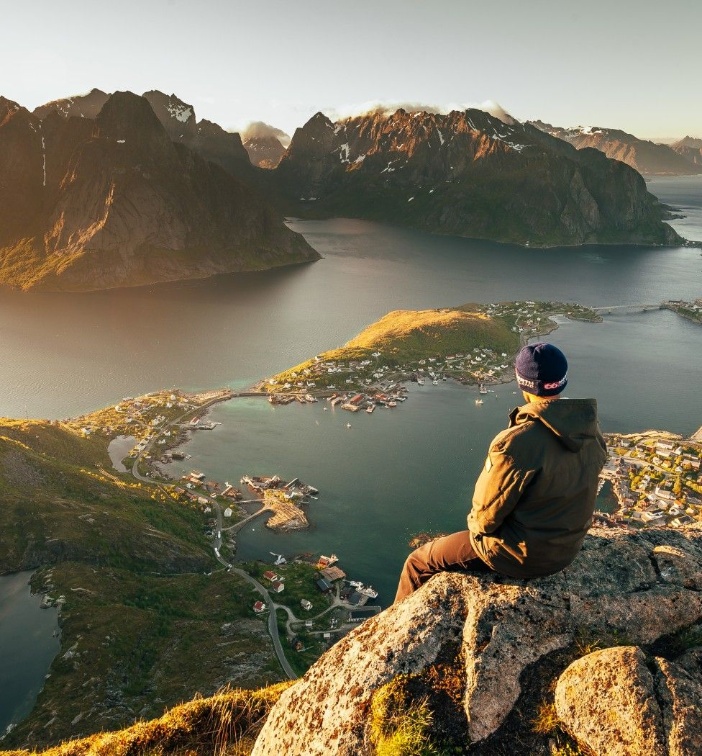} & 
	\includegraphics[height=2.7cm]{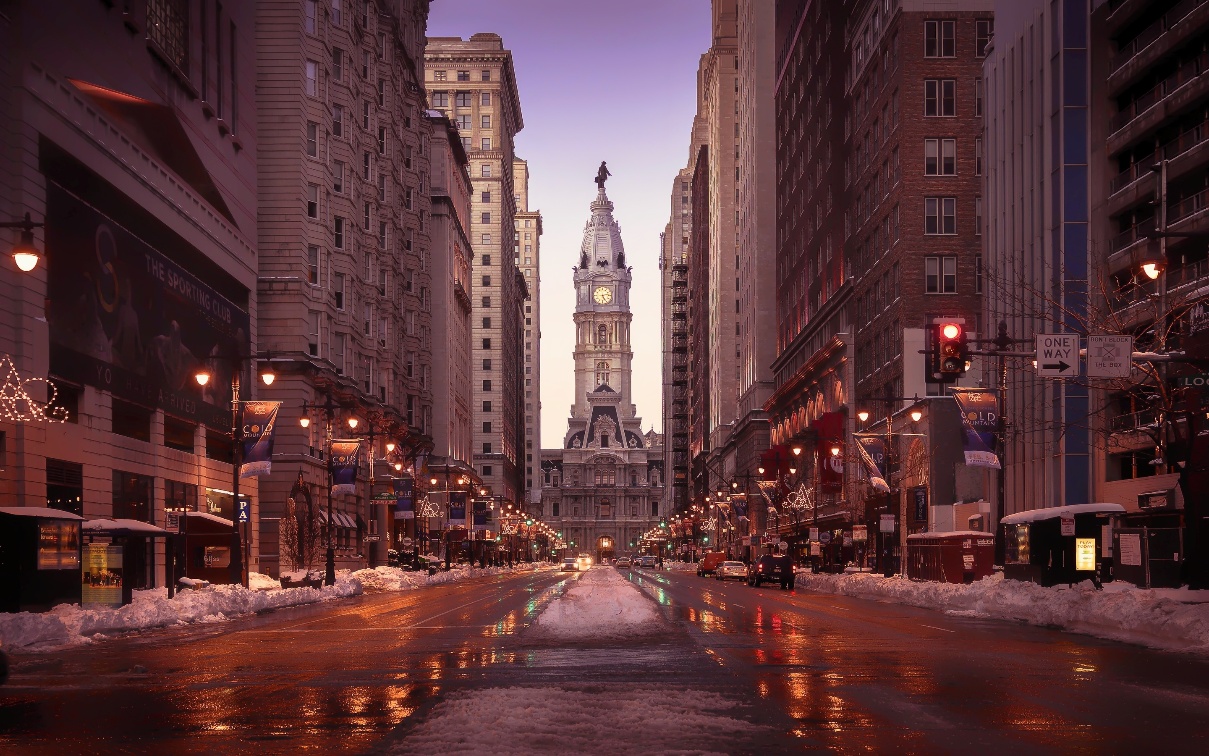} & 
	\includegraphics[trim={70pt 0 77pt 0},clip,height=2.7cm]{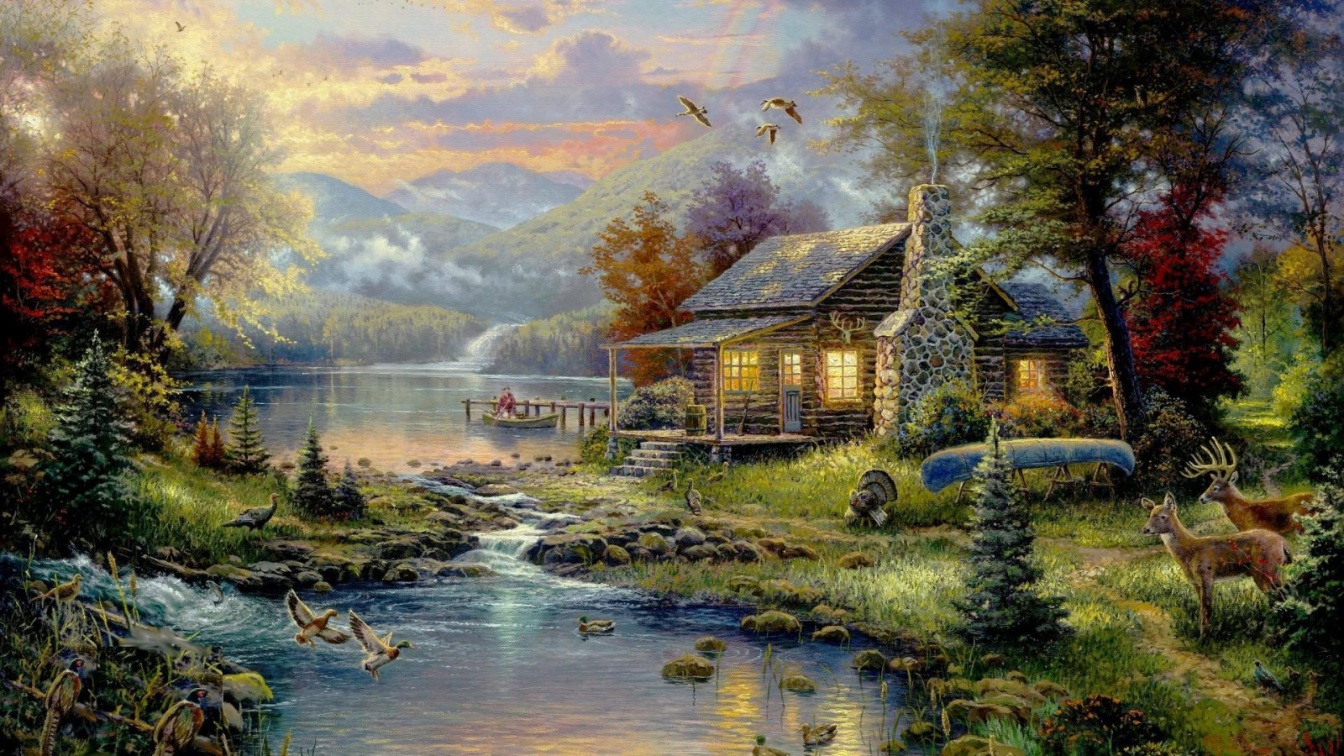} \\
    \includegraphics[trim={0 35pt 0 0},clip,height=2.7cm]{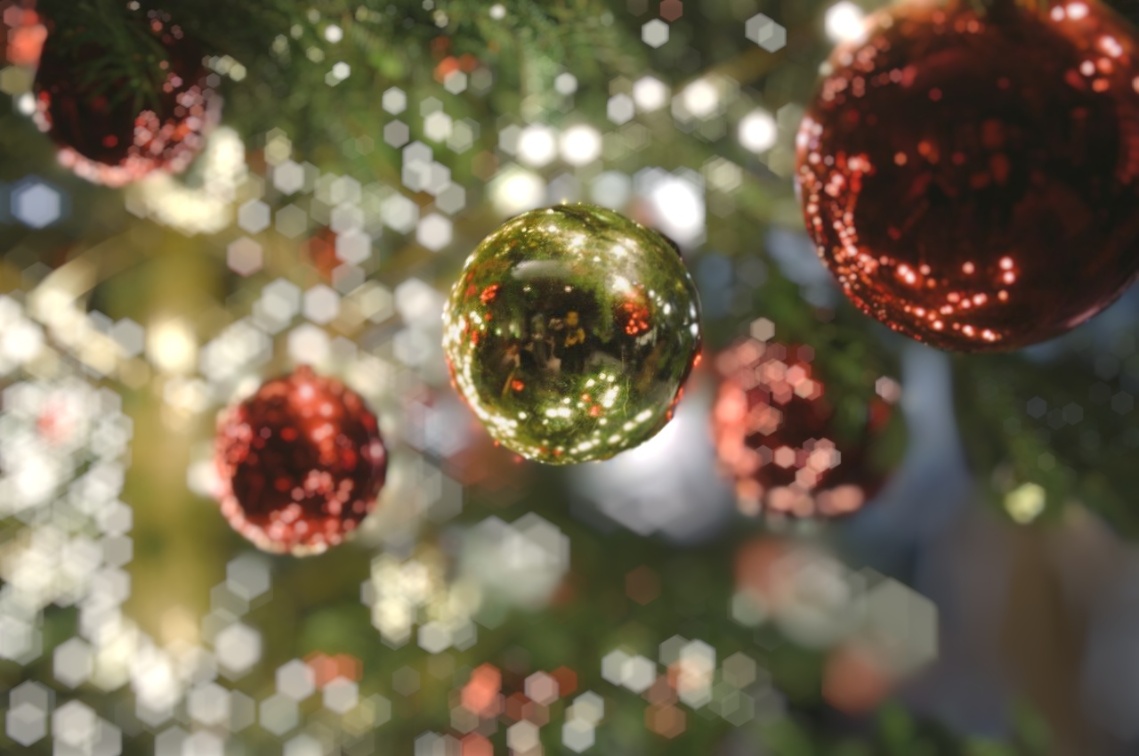} &
    \includegraphics[trim={0 35pt 0 245pt},clip,height=2.7cm]{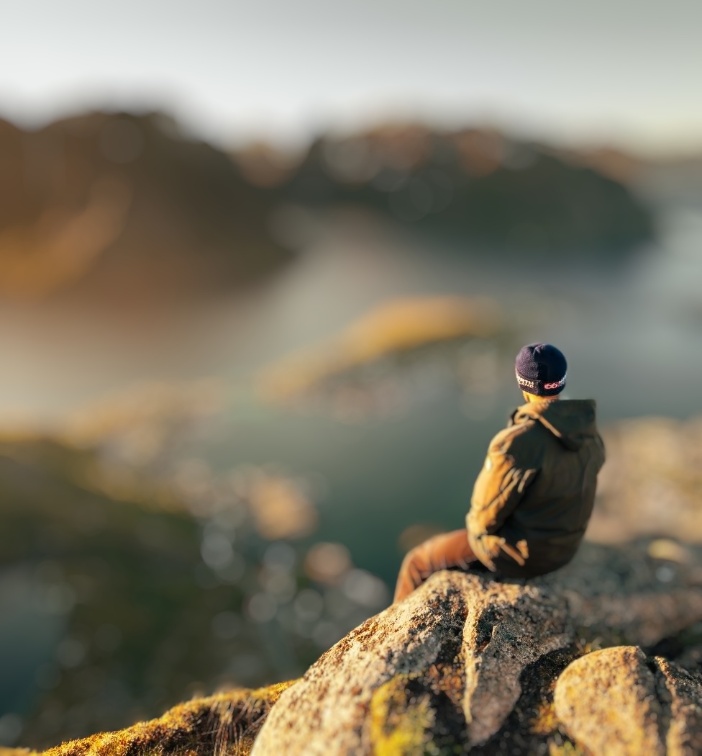} & 
	\includegraphics[height=2.7cm]{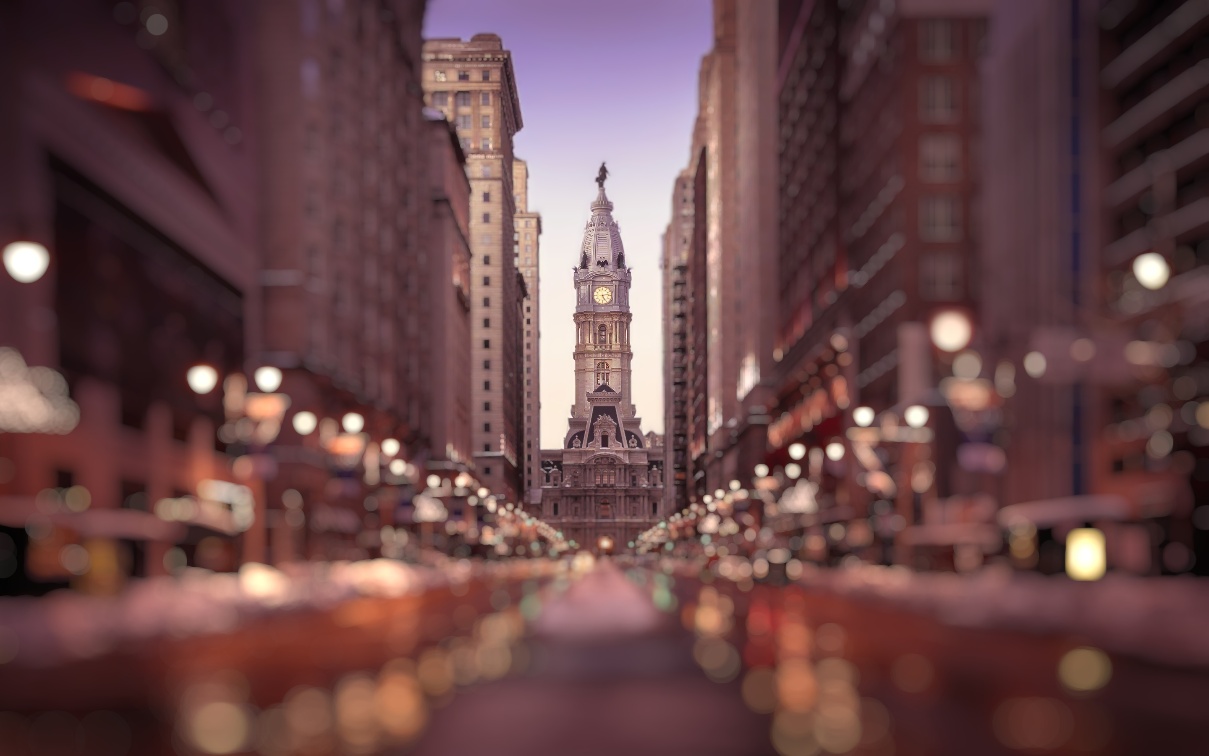} & 
	\includegraphics[trim={70pt 0 77pt 0},clip,height=2.7cm]{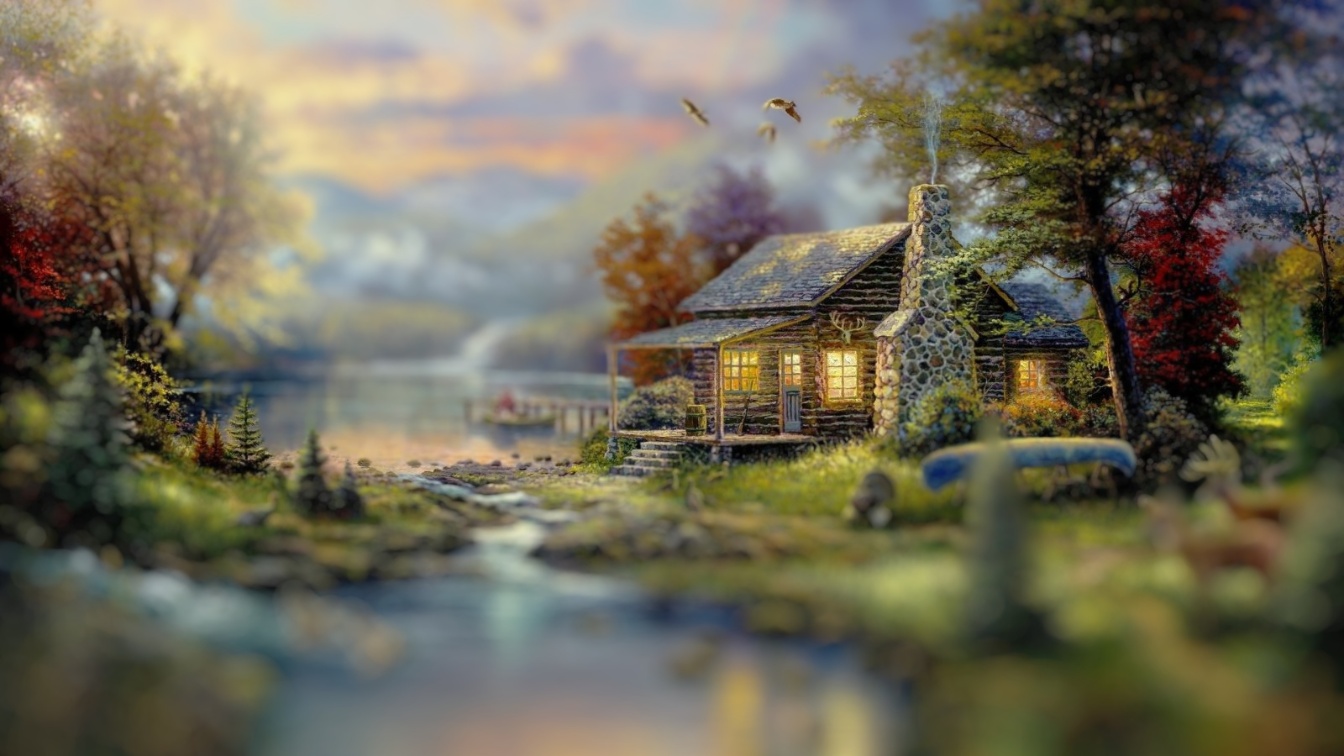} \\
\end{tabular}

\captionof{figure}{
BokehMe creates photo-realistic and highly controllable bokeh effects from high-resolution images and imperfect disparity maps predicted by DPT~\cite{ranftl2021vision}. The first column shows the result with a hexagon aperture shape, and the rest of them use a circular shape.
}
\label{fig:teaser}
\end{center}
}]

\renewcommand{\thefootnote}{\fnsymbol{footnote}} 
\footnotetext[1]{Corresponding author.}

\begin{abstract}
We propose BokehMe, a hybrid bokeh rendering framework that marries a neural renderer with a classical physically motivated renderer. Given a single image and a potentially imperfect disparity map, BokehMe generates high-resolution photo-realistic bokeh effects with adjustable blur size, focal plane, and aperture shape. To this end, we analyze the errors from the classical scattering-based method and derive a formulation to calculate an error map. Based on this formulation, we implement the classical renderer by a scattering-based method and propose a two-stage neural renderer to fix the erroneous areas from the classical renderer. The neural renderer employs a dynamic multi-scale scheme to efficiently handle arbitrary blur sizes, and it is trained to handle imperfect disparity input. Experiments show that our method compares favorably against previous methods on both synthetic image data and real image data with predicted disparity. A user study is further conducted to validate the advantage of our method.  

\end{abstract}

\section{Introduction}
Bokeh effect refers to the way the lens renders the out-of-focus blur in a photograph (Fig.~\ref{fig:teaser}). With different lens designs and configurations, various bokeh styles can be created. For example, the shape of the bokeh ball can be controlled by the aperture.
Classical rendering methods~\cite{busam2019sterefo,luo2020bokeh,wadhwa2018synthetic,zhang2019synthetic} can change bokeh styles easily by controlling the shape and size of the blur kernel. However, they often suffer from 
artifacts at depth discontinuities. Neural rendering methods~\cite{ignatov2020rendering,qian2020bggan,wang2018deeplens} can address this problem well by learning from image statistics, but they have difficulty simulating real bokeh balls and can only produce the bokeh style from the training data. In addition, previous neural rendering methods lack a mechanism to produce large blur size on high-resolution images, because of the fixed receptive field of the neural network and the blur size limit of the training data.

\begin{figure}
    \setlength{\abovecaptionskip}{3pt}
    \setlength{\belowcaptionskip}{-10pt}
    \small
	\centering
	\renewcommand\arraystretch{.2}
    \begin{tabular}{*{4}{c@{\hspace{.2mm}}}}
        \multicolumn{1}{l}{\multirow{3}{*}[28.3pt]{
            \hspace{-10pt}
            \begin{tikzpicture}
            \node[inner sep=0]{\includegraphics[width=0.307\linewidth]{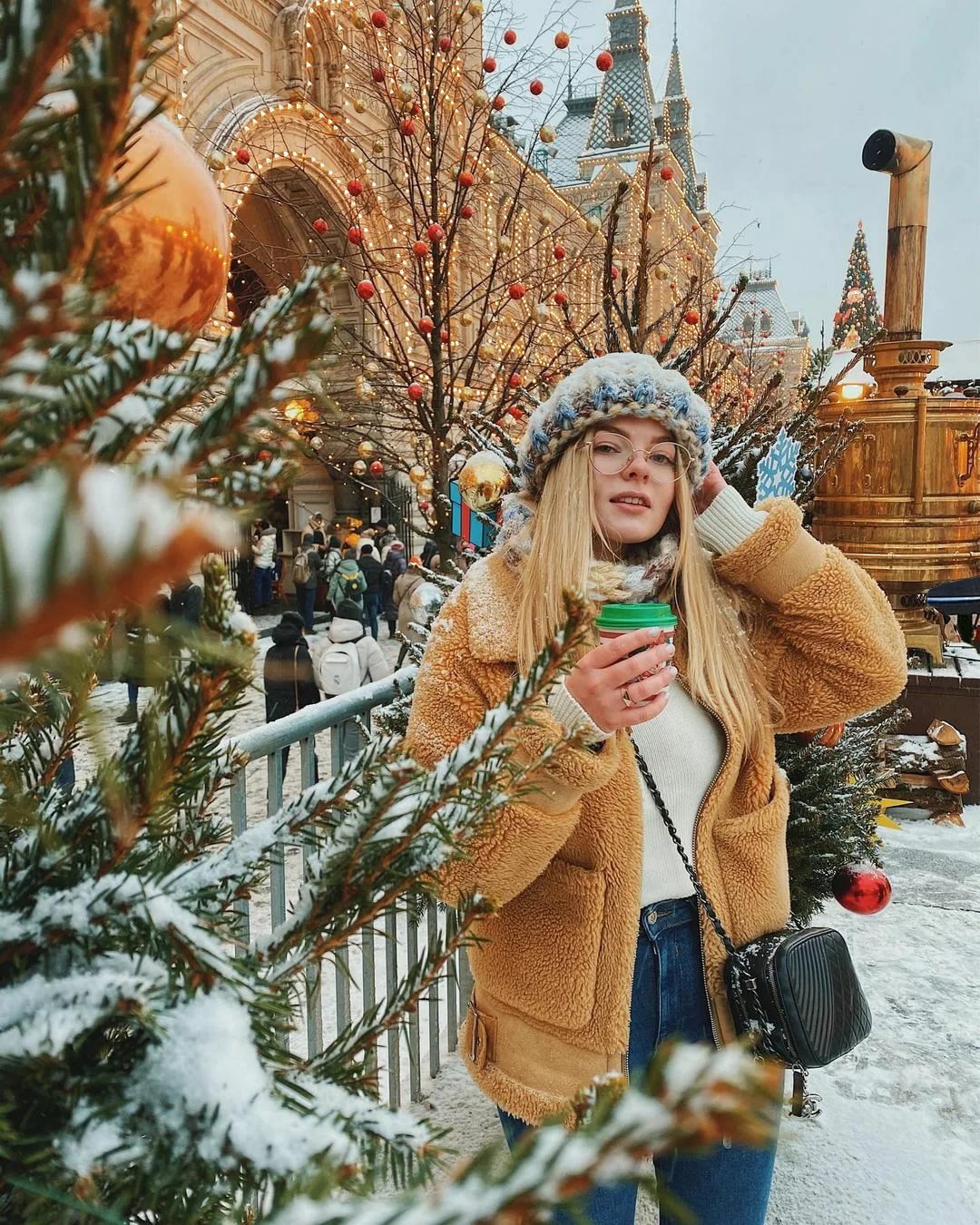}};
            \draw[thick,red] (-0.41,0.86) rectangle ++(0.391,0.22);
            \draw[thick,green] (0.15,0.54) rectangle ++(0.391,0.22);
            \end{tikzpicture}
            \hspace{-9.7pt}
        }} &
        
        \begin{tikzpicture}
        \draw[red,fill=red] (-0.925,-0.52) rectangle (0.925,0.52);
        \node[inner sep=0]{\includegraphics[width=0.218\linewidth]{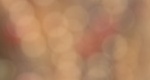}};
        \end{tikzpicture} &
        
        \begin{tikzpicture}
        \draw[red,fill=red] (-0.925,-0.52) rectangle (0.925,0.52);
        \node[inner sep=0]{\includegraphics[width=0.218\linewidth]{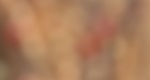}};
        \end{tikzpicture} &
        
        \begin{tikzpicture}
        \draw[red,fill=red] (-0.925,-0.52) rectangle (0.925,0.52);
        \node[inner sep=0]{\includegraphics[width=0.218\linewidth]{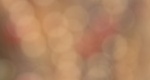}};
        \end{tikzpicture} \\
        
        ~ & 
        
        \begin{tikzpicture}
        \draw[red,fill=red] (-0.925,-0.52) rectangle (0.925,0.52);
        \node[inner sep=0]{\includegraphics[width=0.218\linewidth]{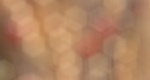}};
        \end{tikzpicture} &
        
        \begin{tikzpicture}
        \draw[red,fill=red] (-0.925,-0.52) rectangle (0.925,0.52);
        \node[inner sep=0]{\includegraphics[width=0.218\linewidth]{crop1_bokeh_neural.jpg}};
        \end{tikzpicture} &
        
        \begin{tikzpicture}
        \draw[red,fill=red] (-0.925,-0.52) rectangle (0.925,0.52);
        \node[inner sep=0]{\includegraphics[width=0.218\linewidth]{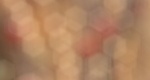}};
        \end{tikzpicture} \\
        
        ~ & 
        
        \begin{tikzpicture}
        \draw[green,fill=green] (-0.925,-0.52) rectangle (0.925,0.52);
        \node[inner sep=0]{\includegraphics[width=0.218\linewidth]{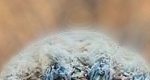}};
        \end{tikzpicture} &
        
        \begin{tikzpicture}
        \draw[green,fill=green] (-0.925,-0.52) rectangle (0.925,0.52);
        \node[inner sep=0]{\includegraphics[width=0.218\linewidth]{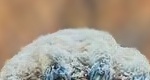}};
        \end{tikzpicture} &
        
        \begin{tikzpicture}
        \draw[green,fill=green] (-0.925,-0.52) rectangle (0.925,0.52);
        \node[inner sep=0]{\includegraphics[width=0.218\linewidth]{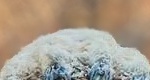}};
        \end{tikzpicture} \\\\
        \noalign{\vskip 0.4mm}
        \hspace{-5pt} All-in-Focus & Ours (CR) & Ours (NR) & Ours \\
    
    \end{tabular}
    \caption{BokehMe combines a classical renderer (CR) and a neural renderer (NR) to create bokeh effects with stunning bokeh balls and adjustable aperture shapes (row 1: circle; row 2: hexagon).}
	\label{fig:effect}
\end{figure}

To produce artifact-free and highly controllable bokeh effects, we propose a novel hybrid framework, termed BokehMe, which makes the best of the two worlds by fusing the results from a classical renderer and a neural renderer (Fig.~\ref{fig:effect}).
We use the scattering-based method~\cite{wadhwa2018synthetic} as our classical renderer.
To determine where this method may render noticeable boundary artifacts, we model the lens system and conduct a comprehensive analysis of the error between scattering-based rendering and real rendering. A soft but tight error map is derived to identify regions with boundary artifacts. Using the error map to replace the artifact region with the neural rendering result, we are able to preserve the bokeh style from the classical renderer without apparent visual artifacts.
For the neural renderer, to break the blur size limit, we decompose it to two sub-networks: adaptive rendering network (ARNet) and iterative upsampling network (IUNet). In ARNet, we resize the input images adaptively and generate a bokeh image in low resolution. Then, IUNet is used to upsample the low-resolution bokeh image iteratively guided by the initial high-resolution input images.
As a result, our neural renderer can handle arbitrarily large blur sizes. 



Our main contributions are summarized as follows.
\begin{itemize}
    \item[$\bullet$] We propose a novel framework, which combines a classical renderer and a neural renderer for photo-realistic and highly controllable bokeh rendering.
    \item[$\bullet$] We analyze the lens system and propose an error map formulation to effectively fuse the classical rendering and the neural rendering.
    \item[$\bullet$] We propose a two-stage neural renderer which uses adaptive resizing and iterative upsampling to handle arbitrary blur sizes for high-resolution images, and it is robust to potentially imperfect disparity input.
\end{itemize}

In addition, due to the lack of test data in the field of controllable bokeh rendering,
we contribute a new benchmark: BLB, synthesized by Blender 2.93~\cite{blender}, together with EBB400, processed from EBB!~\cite{ignatov2020rendering}.
Since the evaluation of bokeh effects is subjective, we also conduct a user study on images captured by iPhone 12. Extensive results show that BokehMe can render images that appear physically sound and maintain the diversity of the bokeh style.

\section{Related Work}
\noindent\textbf{Classical Rendering.} Classical rendering can be classified into two categories: object space methods and image space ones. Object space methods~\cite{abadie2018advances,lee2010real,wu2013rendering,yu2010real}, based on ray tracing, render exact results. However, most are time-consuming and require complete 3D scene information, resulting in poor practicality. Compared with object space methods, image space ones~\cite{barron2015fast,bertalmio2004real,hach2015cinematic,soler2009fourier,yang2016virtual} only 
require a single image and its corresponding depth map, which are easier to implement. 
In recent years, more and more methods~\cite{busam2019sterefo,luo2020bokeh,peng2021interactive,shen2016automatic,shen2016deep,wadhwa2018synthetic,xian2021ranking,zhang2019synthetic} combine different modules, such as depth estimation, semantic segmentation, and classical rendering, to construct an automatic rendering system.
To prevent the color of background from bleeding into foreground, most methods decompose the image to multiple layers conditioned on the estimated depth map, and execute rendering from back to front.

Despite the fact that classical rendering is flexible, this paradigm suffers from artifacts at depth discontinuities, especially when the focal plane targets background.

\vspace{3pt}
\noindent\textbf{Neural Rendering.} To improve efficiency and avoid boundary artifacts, many recent works use neural networks to simulate the rendering process. For example, Nalbach \etal~\cite{nalbach2017deep} and Xiao \etal~\cite{xiao2018deepfocus} train networks to produce a bokeh effect from an all-in-focus image and its corresponding perfect depth map. By training on the synthetic data created by OpenGL shaders and Unity Engines~\cite{unity}, boundary artifacts can be effectively alleviated.
However, perfect depth maps are not always easy to obtain in the real world.
Wang \etal~\cite{wang2018deeplens} thus propose an automatic rendering system comprised of depth prediction, lens blur, and guided upsampling to generate high-resolution depth-of-field (DoF) images from a single image.
Besides, encoder-decoder networks~\cite{dutta2021stacked,ignatov2019aim,ignatov2020aim,ignatov2020rendering,qian2020bggan}, that map all-in-focus images into shallow DoF images in an end-to-end manner, have also been studied recently. Unlike aforementioned methods, Xu \etal~\cite{xu2018rendering} focus on fully automatic portrait rendering. They use recurrent filters~\cite{liu2016learning} to approximate the conditional random field-based rendering method and achieve a significant speed improvement.

However, the main problem of neural rendering is lack of controllability. For a trained neural network, the bokeh style cannot be changed and the blur range is limited. In addition, bokeh balls produced by the network are not real as the network tends to learn a simple fuzzy effect.


\begin{figure}
\setlength{\abovecaptionskip}{-5pt}
\setlength{\belowcaptionskip}{-5pt}
\begin{center}
\includegraphics[width=0.95\linewidth]{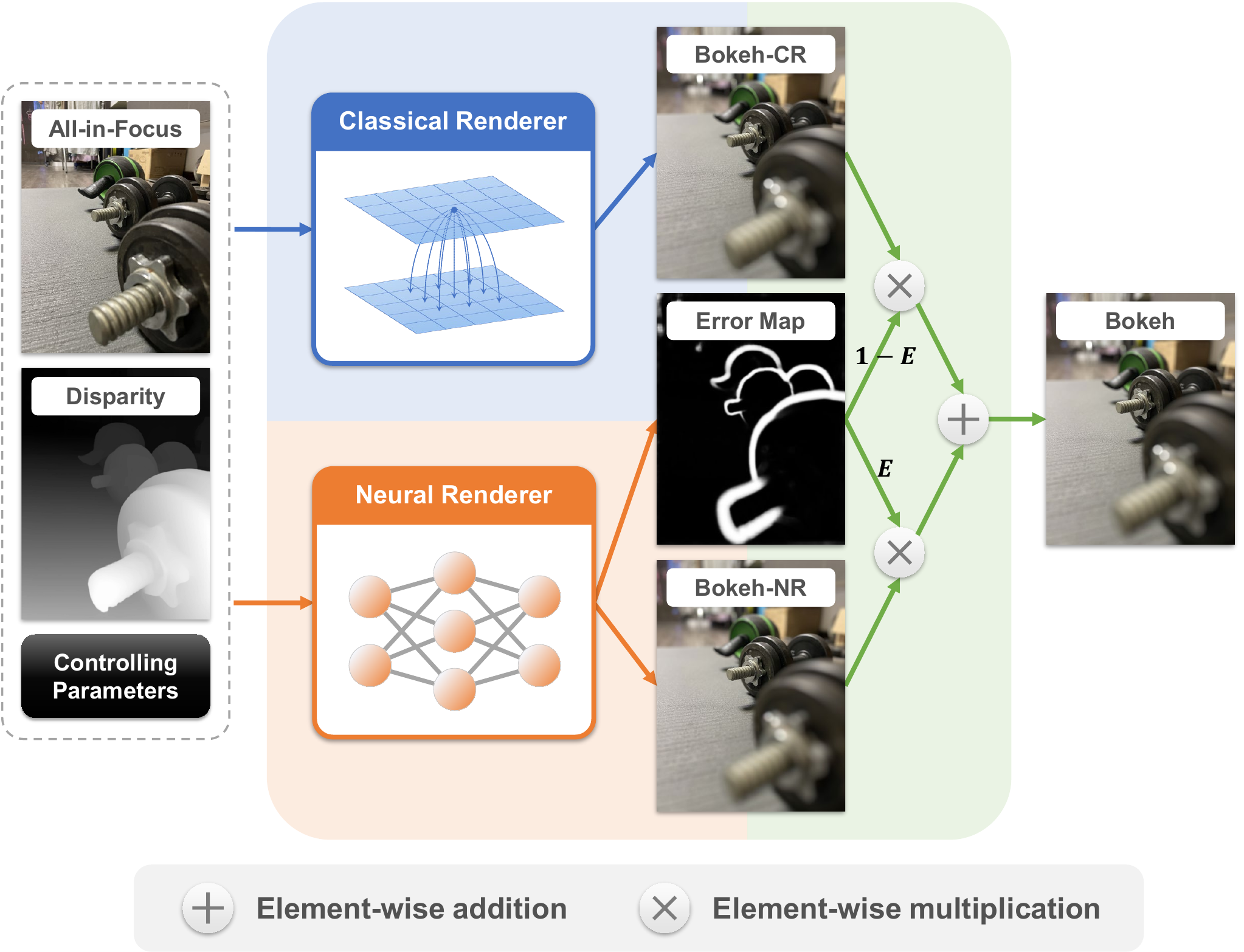}
\end{center}
\caption{
Framework of BokehMe. The bokeh image is obtained by fusing the outputs of a classical renderer and a neural renderer.}
\label{fig:framework}
\end{figure}

\section{BokehMe: A Hybrid Rendering Framework}
\label{sec:bokehme}
As shown in Fig.~\ref{fig:framework}, our framework generates a bokeh image $B$ from an all-in-focus image $I$, a disparity map $D$, and controlling parameters via two renderers: a classical renderer and a neural renderer. Their rendered results are fused based on an error map $E$ that identifies the potentially erroneous areas from the classical renderer. The controlling parameters include blur parameter $K$, refocused disparity $d_f$, gamma value $\gamma$, and some parameters about the bokeh style, \eg, aperture shape. Specifically, $K$ reflects the blur amount of the whole image. $d_f$ determines the disparity (inverse depth) of the focal plane. $\gamma$, used in gamma correction, controls the brightness and salience of bokeh balls.

\subsection{Classical Renderer and Error Analysis}
\label{sec:pr}
\noindent\textbf{Classical Renderer.} 
We expect the classical renderer to focus on rendering realistic bokeh effects in depth-continuous areas. After comparing different methods, we find pixel-wise rendering methods based on scattering~\cite{peng2021interactive,wadhwa2018synthetic} have relatively small error in these areas despite causing serious color-bleeding artifacts at depth discontinuities.
The core idea of this method is scattering each pixel to its neighbor areas where the distance between them is less than the blur radius of the pixel. As discussed in~\cite{wadhwa2018synthetic,yang2016virtual}, given the disparity $d$ of a pixel, its blur radius can be calculated by
\begin{equation}\label{eq:radius}
    r=K\,\lvert d-d_f \rvert\,.
\end{equation}
We implement the algorithm with CuPy package to achieve a significant parallel speedup (refer to the supplementary material). Since the transformation from scene irradiance to image intensity is nonlinear~\cite{yang2016virtual}, an additional gamma correction~\cite{lin2011revisiting} is applied before and after the rendering. 

\begin{figure}
\setlength{\abovecaptionskip}{-5pt}
\setlength{\belowcaptionskip}{-5pt}
\begin{center}
\includegraphics[width=0.95\linewidth]{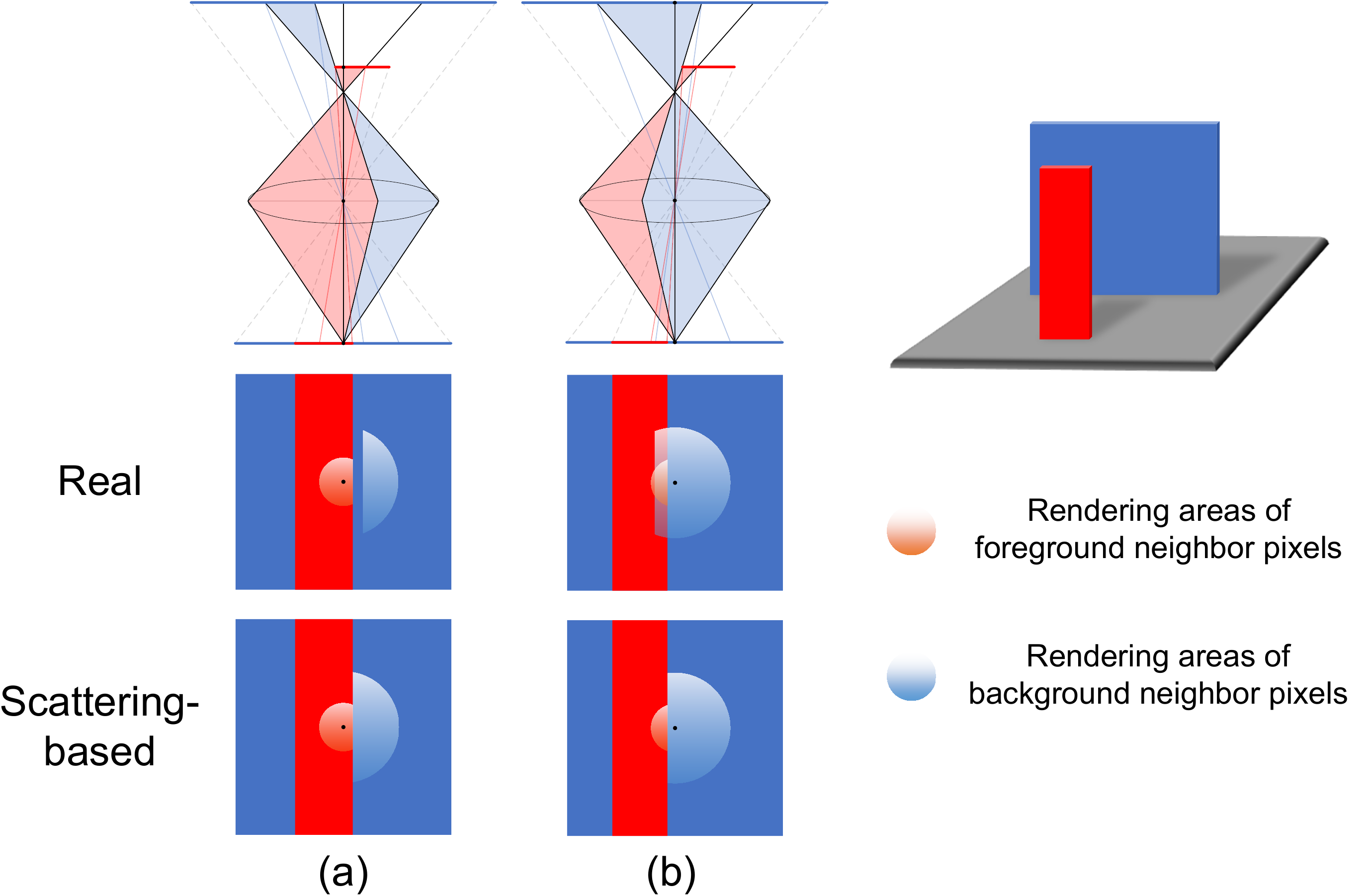}
\end{center}
\caption{
Comparison of real rendering and scattering-based rendering at depth discontinuities. 
Rendered result of the center pixel (the black dot) is the integration of its neighbor pixels on red gradient foreground plane and those on blue gradient background plane. 
}
\label{fig:lens_system}
\end{figure}

\vspace{3pt}
\noindent\textbf{Lens System.} To understand why scattering-based methods cause error at depth discontinuities, we model a virtual lens system. For a simple scenario (Fig.~\ref{fig:lens_system}) where two objects exist in space, we derive $8$ rendering cases at depth discontinuities ($2$ cases are shown here while the others are shown in the supplementary material). 
Taking the center pixel (the black dot) as an example, only neighbor pixels on red gradient foreground plane and those on blue gradient background plane can pass to the center pixel. Apparently, the scattering-based rendering is different from the real one.

\vspace{3pt}
\noindent\textbf{Initial Error Map.} We aim to obtain an error map to identify areas rendered incorrectly by the classical renderer. 
Later, we will train a neural network to predict the error map formulated in this section. Let $E^*$ denote the target error map.
Since only regions within the scattering radius from the depth boundary may have significant difference from the real rendering, $E^*$ can be conservatively formulated as the spatially variant dilation of the depth boundary, and the dilation size depends on the maximum blur radius of the pixels located on both sides of the depth boundary. 
Take the scenario in Fig.~\ref{fig:lens_system} as an example, the $i$-th element of $E^*$ can be defined by
\begin{equation}\label{eq:var1}
    E^*_i = \mathbbm{1}\big(\alpha_i<1\big)\,,\quad \alpha_i = \frac{l_{ii^\prime}}{\max{(r_i,r_{i^\prime})}}\,,
\end{equation}
where $\alpha_i$ can be treated as a variable of $E^*_i$. $i^\prime$ is the index to the nearest pixel of the $i$-th pixel in the other depth plane. $l_{ii^\prime}$ is the distance between the two pixels. $r_i$ and $r_{i^\prime}$ are the blur radii of the corresponding pixels. 

\vspace{3pt}
\noindent\textbf{Improved Error Map.} Considering the fact that the classical renderer generates high-quality results in depth-continuous regions with controllable bokeh style, we would like to appropriately narrow and soften the initial error map to preserve more of the bokeh result from the classical renderer without obvious artifacts in fusion boundary.

\begin{figure}
    \setlength{\abovecaptionskip}{5pt}
    \setlength{\belowcaptionskip}{-5pt}
    \footnotesize
	\centering
	\renewcommand\arraystretch{.5}
	\begin{tabular}{*{3}{c@{\hspace{.6mm}}}}
        \includegraphics[width=0.31\linewidth]{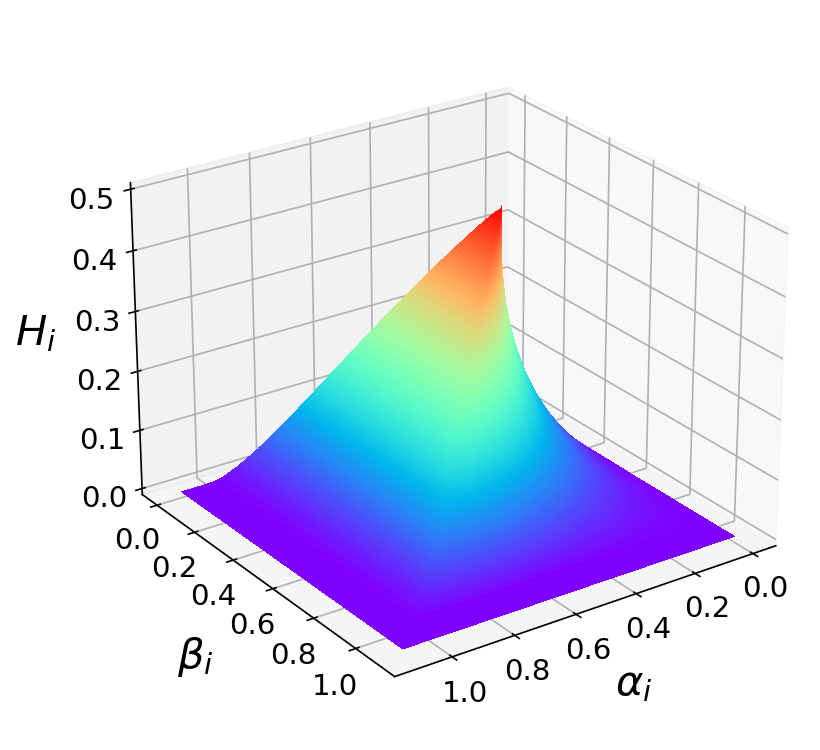} &
        \includegraphics[width=0.31\linewidth]{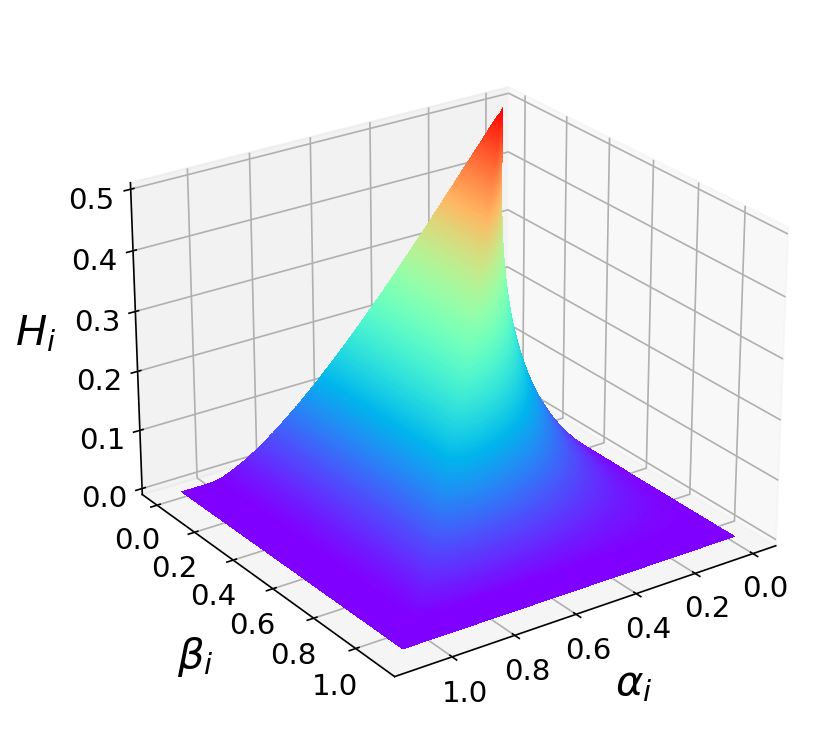} &
        \includegraphics[width=0.31\linewidth]{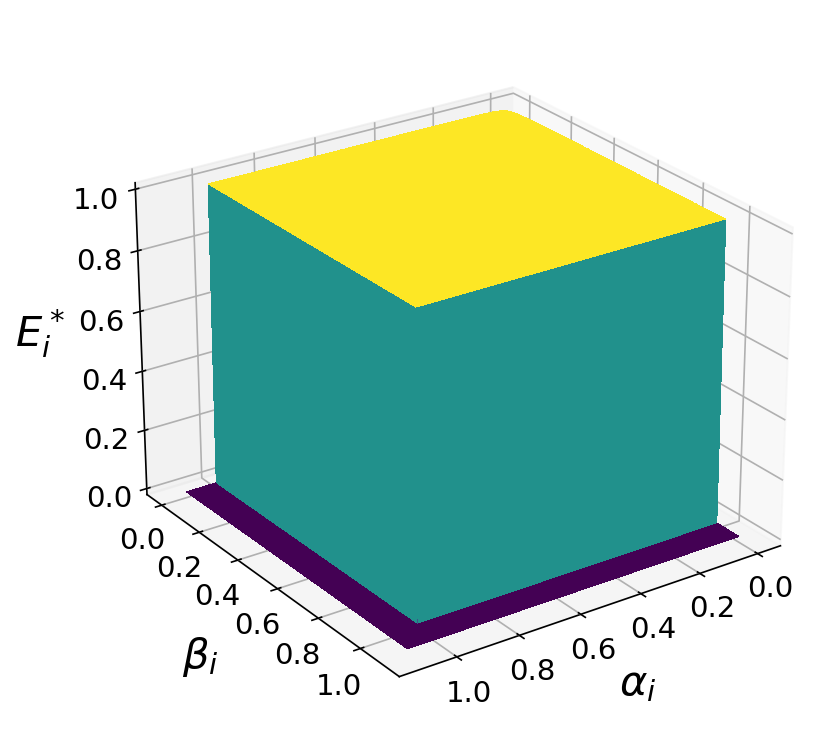} \\
        Color Diff. (case 1) & Color Diff. (case 2) & Initial Error Map \\
        \includegraphics[width=0.31\linewidth]{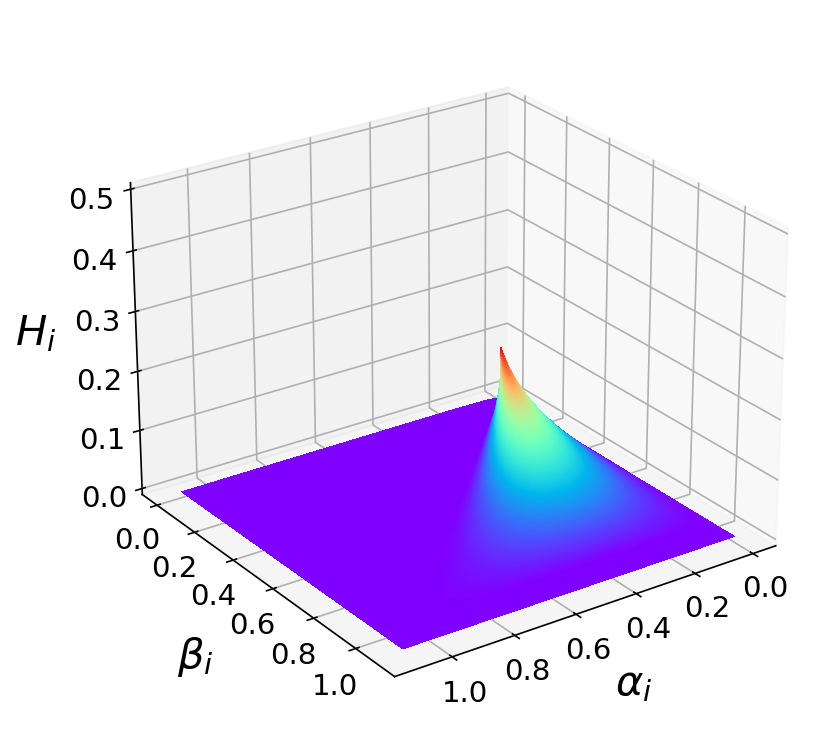} &
        \includegraphics[width=0.31\linewidth]{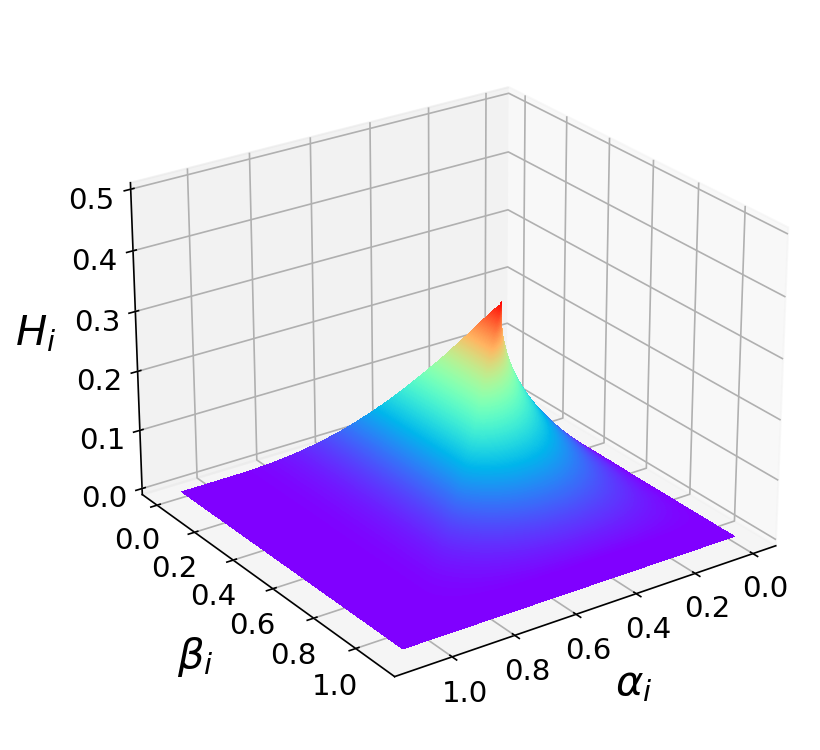} &
        \includegraphics[width=0.31\linewidth]{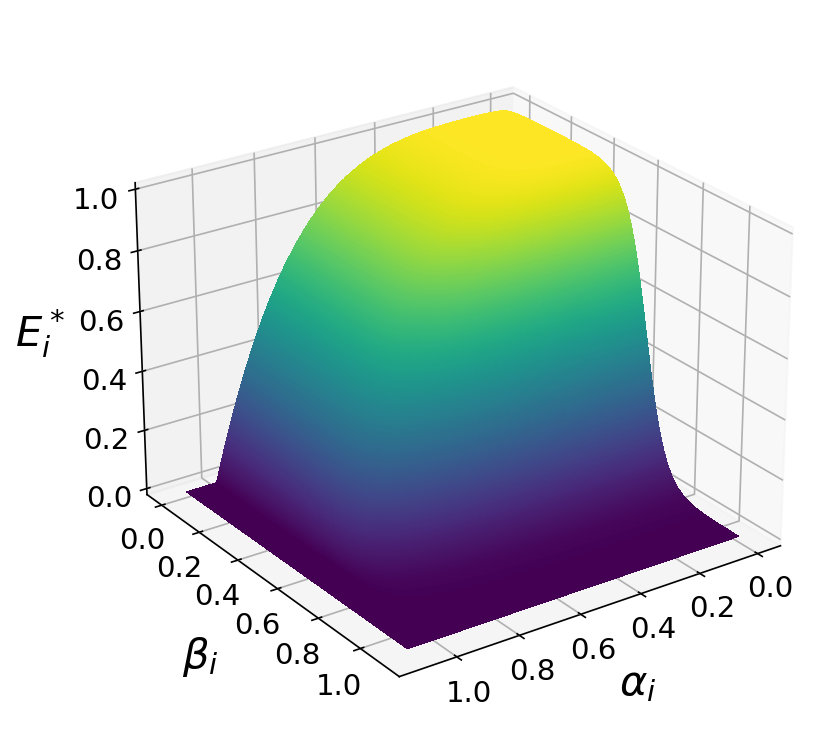} \\
		Color Diff. (case 3) & Color Diff. (case 4) & Improved Error Map \\
	\end{tabular}
	\caption{Column 1, 2: graphs of the color difference $H_i$ between real rendering and scattering-based rendering in $4$ cases. Column 3: graphs of the error map $E^*_i$ \wrt the erroneous areas from the classical renderer. The improved error map is softer and tighter than the initial one, and covers the color difference on the whole. 
	}
	\label{fig:error_plot}
\end{figure}

\begin{figure}
    \setlength{\abovecaptionskip}{5pt}
    \small
	\centering
	\renewcommand\arraystretch{.5}
	\begin{tabular}{*{4}{c@{\hspace{.6mm}}}}
        \multicolumn{2}{c}{
            \hspace{-4pt}
            \begin{tikzpicture}
            \node[inner sep=0]{\includegraphics[trim={0 80pt 0 100pt},clip,width=0.475\linewidth]{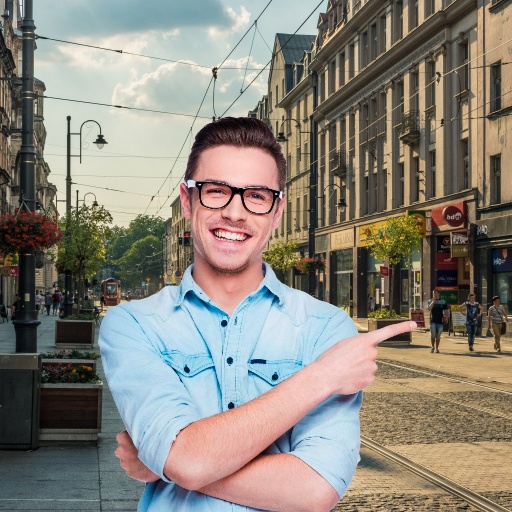}};
            \draw[thick,green] (0.7,-0.75) rectangle ++(0.65,0.4);
            \end{tikzpicture}
            \hspace{-8pt}
        } &
        \multicolumn{2}{c}{
            \hspace{-9.5pt}
            \begin{tikzpicture}
            \node[inner sep=0]{\includegraphics[trim={0 80pt 0 100pt},clip,width=0.475\linewidth]{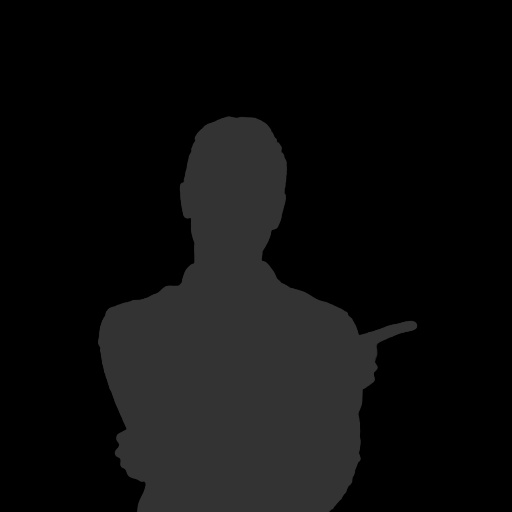}};
            \draw[thick,green] (0.7,-0.75) rectangle ++(0.65,0.4);
            \end{tikzpicture}
            \hspace{-8pt}
        } \\
        \includegraphics[width=0.235\linewidth]{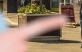} &
        \includegraphics[width=0.235\linewidth]{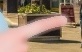} &
        \includegraphics[width=0.235\linewidth]{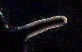} &
        \includegraphics[width=0.235\linewidth]{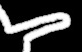} \\
        \includegraphics[width=0.235\linewidth]{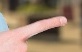} &
        \includegraphics[width=0.235\linewidth]{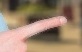} &
        \includegraphics[width=0.235\linewidth]{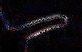} &
        \includegraphics[width=0.235\linewidth]{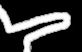} \\
        \includegraphics[width=0.235\linewidth]{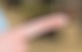} &
        \includegraphics[width=0.235\linewidth]{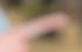} &
        \includegraphics[width=0.235\linewidth]{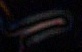} &
        \includegraphics[width=0.235\linewidth]{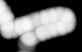} \\
        \includegraphics[width=0.235\linewidth]{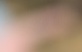} &
        \includegraphics[width=0.235\linewidth]{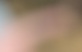} &
        \includegraphics[width=0.235\linewidth]{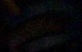} &
        \includegraphics[width=0.235\linewidth]{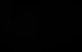} \\
        \noalign{\vskip 0.8mm}
		Real & Scattering & Color Diff. & Error Map \\
	\end{tabular}
	\caption{In this example, the disparities of background and foreground are fixed to $0$ and $0.2$, respectively. The refocused disparity is set to $0$, $0.2$, $0.5$ and $1$ from row 2 to row 5, and the variable $\beta_i$ for each case can be calculated by Eq.~\ref{eq:var2}, \ie, $0$, $0$, $0.6$ and $0.8$. One can see that the color difference between real rendering and scattering-based rendering ``fades out'' with the increase of $\beta_i$, and our improved error map can cover the color difference on the whole, which is consistent with the observation in Fig~\ref{fig:error_plot}.
	}
	\label{fig:error}
\end{figure}

Through the theoretic and numerical analysis shown in the supplementary material,
we derive that 
for each pixel, the color difference between the scattering-based rendering result and the real rendering result is
\begin{equation}
    H_i=k_i\,\lvert c_i-c_{i^\prime}\rvert\,,\quad k_i=f(\alpha_i,\beta_i)\,,
\end{equation}
where $c_i$ and $c_{i^\prime}$ are the colors of the $i$-th pixel and the $i^\prime$-th pixel before rendering.
$k_i$ is a function of two variables $\alpha_i$ and $\beta_i$. $\alpha_i$ has been defined in Eq.~\ref{eq:var1} while $\beta_i$ takes the form
\begin{equation}\label{eq:var2}
    \beta_i = \frac{\min{(r_i,r_{i^\prime})}}{\max{(r_i,r_{i^\prime})}}\,,
\end{equation}
which represents the ratio of the smaller and the larger blur radius of the two pixels. $k_i$ varies with the refocused disparity and the shortest distance between the processing pixel and the depth boundary. For clarity, we assume $\lvert c_i-c_{i^\prime}\rvert=1$ and draw the graphs of $H_i$ in the first two columns of Fig.~\ref{fig:error_plot}. Based on the observation that $H_i$ is reduced with the increase of $\alpha_i$ and $\beta_i$, we heuristically rewrite Eq.~\ref{eq:var1} to
\begin{equation}\label{eq:improved_E}
    E^*_i = \max{\big(0, 1-{\alpha_i}^{\delta_1}\big)} \cdot \mathbbm{1}\big(\beta_i<\delta_2\big)\,,
\end{equation}
where $\delta_1$ and $\delta_2$ are two hyperparameters. This formula will be equivalent to Eq.~\ref{eq:var1} if setting $\delta_1=\infty$ and $\delta_2=1$. Note that in our implementation, we replace the second indicator function term with a smooth one, \ie, $0.5+0.5\,\tanh{(10\,(\delta_2-\beta_i))}$. 
After comparing the model trained with different hyperparameters (in the supplementary material), we empirically set $\delta_1=4$ and $\delta_2=\frac{2}{3}$. We also show the graphs of the initial $E^*_i$ (Eq.~\ref{eq:var1}) and the improved $E^*_i$ (Eq.~\ref{eq:improved_E}) in the last column of Fig.~\ref{fig:error_plot}. Note that as $0\leq\beta_i\leq1$, we define $E^*_i=0$ if $\beta_i>1$. One can observe that the improved $E^*_i$ is softer and tighter than the initial one, and still covers the area with large color difference. An additional practical example is shown in Fig.~\ref{fig:error}. 

\subsection{Neural Renderer and Model Training}
\label{sec:nr}
\begin{figure}
\setlength{\abovecaptionskip}{-5pt}
\setlength{\belowcaptionskip}{-5pt}
\begin{center}
\includegraphics[width=\linewidth]{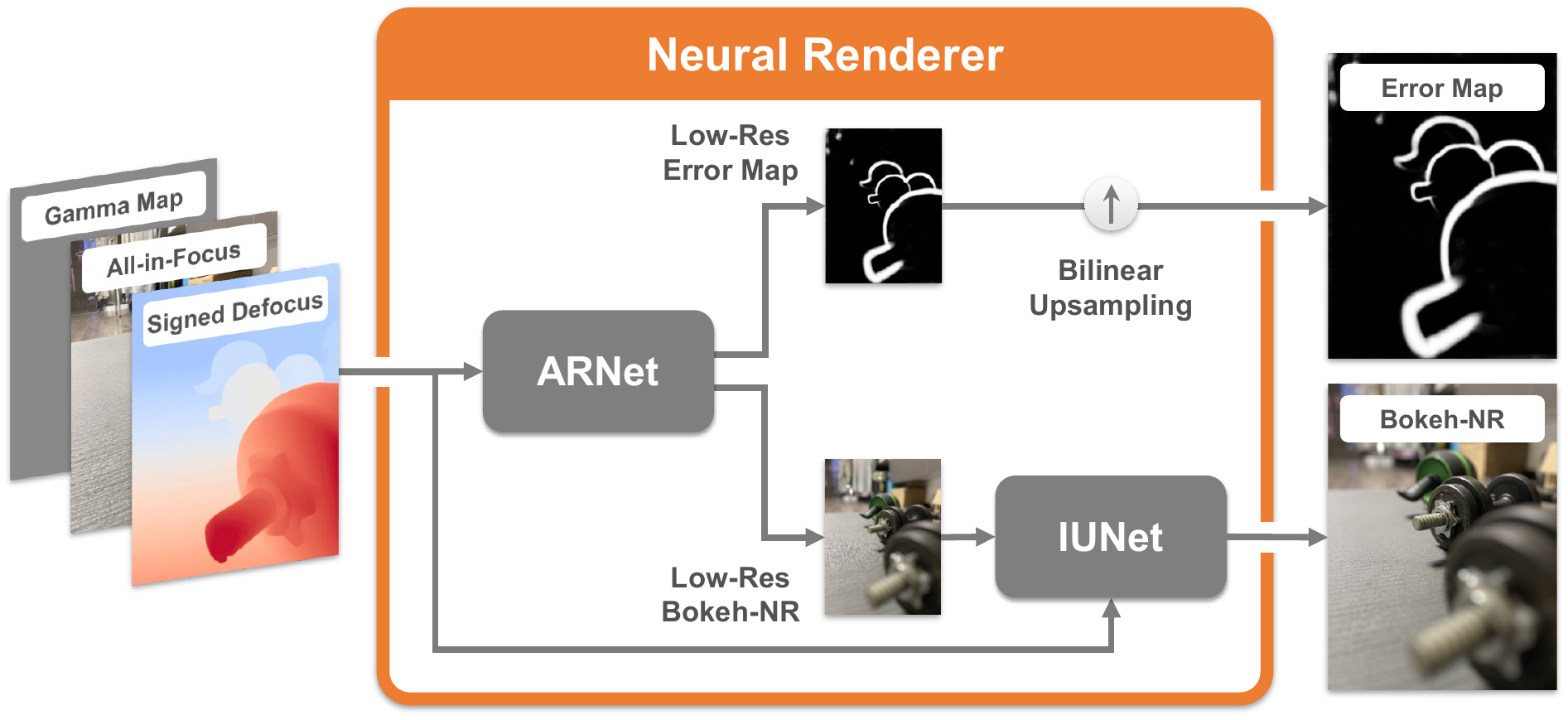}
\end{center}
\caption{Architecture of the neural renderer. ARNet first estimates a low-resolution bokeh image and a low-resolution error map. Then, the error map is restored to original resolution by bilinear upsampling, while the bokeh image is upsampled by IUNet.}
\label{fig:neural_renderer}
\end{figure}

To handle the rendering at depth discontinuities and overcome the limitations of the blur range, we propose a neural renderer consisting of two sub-networks: ARNet and IUNet (Fig.~\ref{fig:neural_renderer}). To simplify the input of the neural renderer, we define a signed defocus map $S$ based on Eq.~\ref{eq:radius}:
\begin{equation}\label{eq:defocus}
    S = K(D-d_f)\,,
\end{equation}
which encodes the information about the depth relationship and the spatially variant blur radius. To match the gamma correction in the classical renderer, we use a map filled with the normalized gamma value as an additional input. 

\begin{figure}
\setlength{\abovecaptionskip}{-5pt}
\setlength{\belowcaptionskip}{-5pt}
\begin{center}
\includegraphics[width=\linewidth]{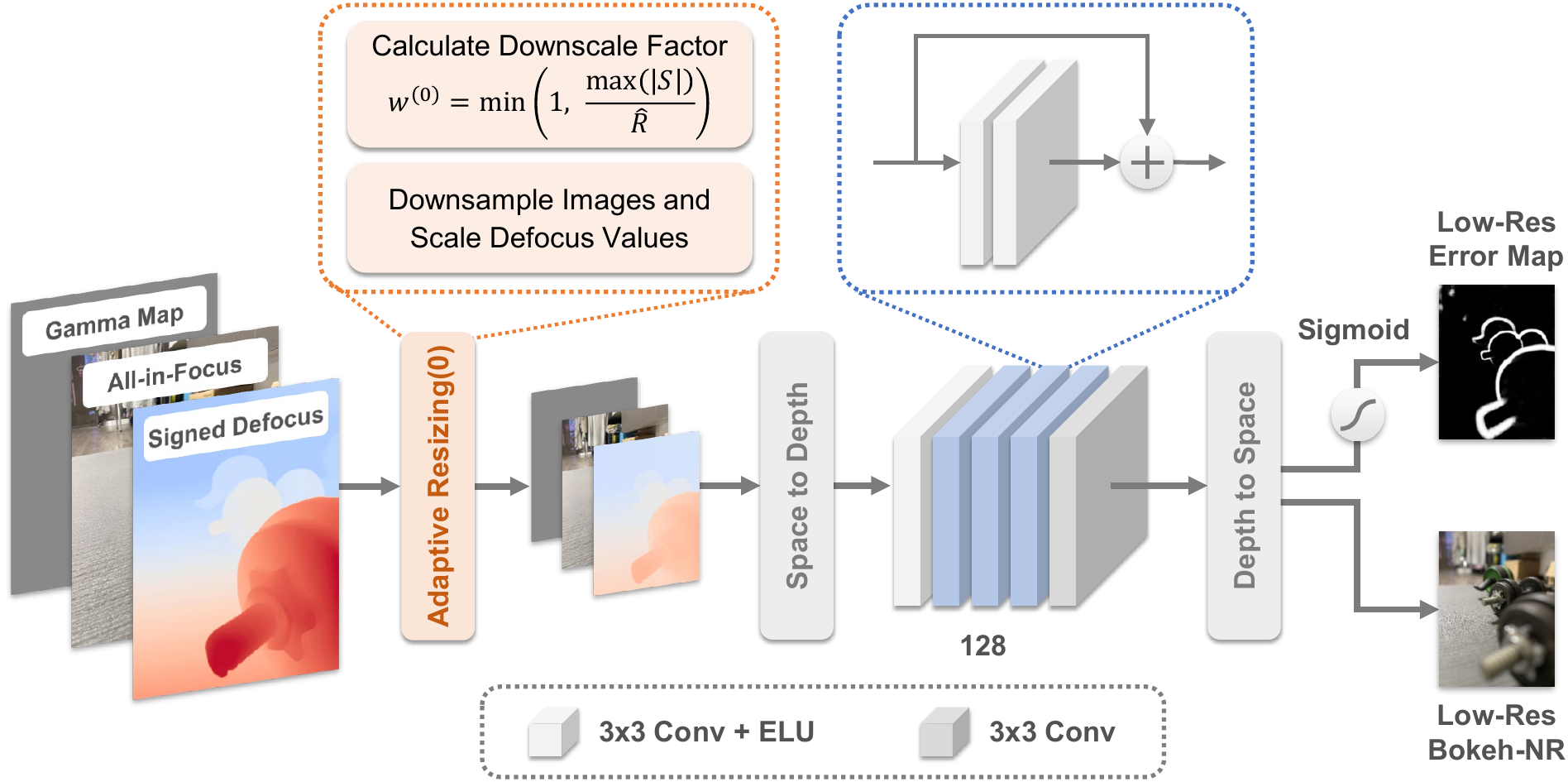}
\end{center}
\caption{Architecture of ARNet. The adaptive resizing layer downsamples the input images and reduces the numerical range of the signed defocus map to ensure that the defocus values are in the acceptable range of the neural network without decreasing the blur amount of the whole image.}
\label{fig:arnet}
\end{figure}

\vspace{3pt}  
\noindent\textbf{ARNet} resizes the input images adaptively, and outputs an error map and a bokeh image $B^{\,lr}_{nr}$ in low resolution (Fig.~\ref{fig:arnet}). The adaptive resizing layer consists of two steps. The first step is to calculate the downscale factor
\begin{equation}
    w^{(0)} = \min{\bigg(1,\,\frac{\max{(\lvert S \rvert)}}{\hat{R}} \bigg)}\,,
\end{equation}
where $\max{(\lvert S \rvert)}$ corresponds to the maximum blur radius of the whole image. $\hat{R}$ is the maximum blur radius we set for the neural network. The second step is to downsample all images and reduce the numerical range of the signed defocus map by the ratio of $w^{(0)}$. The middle part of the network is lightweight and replaceable. We use the same architecture as DeepFocus (fast version)~\cite{xiao2018deepfocus} in this work.

\begin{figure}
\setlength{\abovecaptionskip}{-5pt}
\setlength{\belowcaptionskip}{-5pt}
\begin{center}
\includegraphics[width=\linewidth]{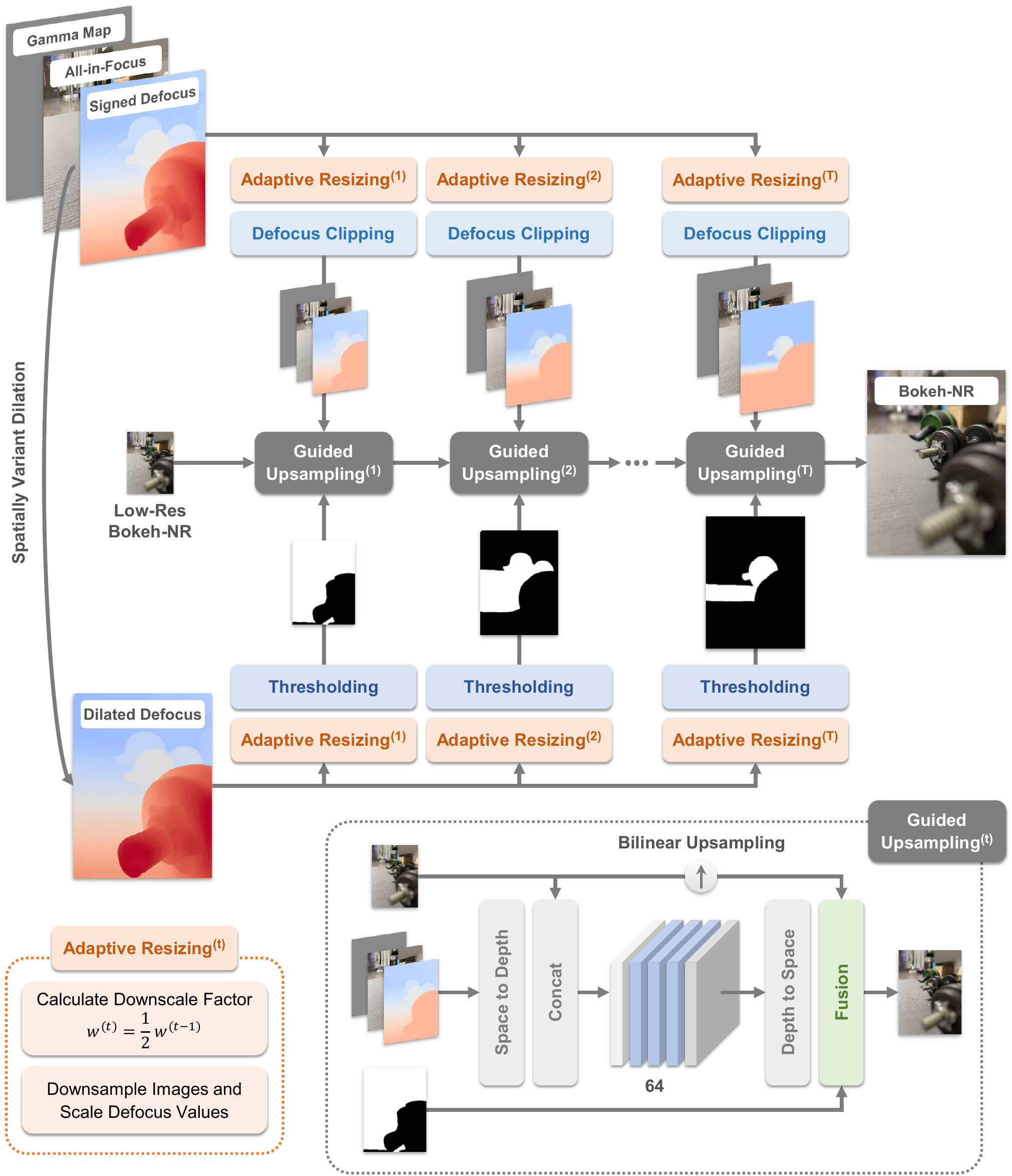}
\end{center}
\caption{Architecture of IUNet. 
The low-resolution bokeh image will be upsampled iteratively to generate a high-quality high-resolution bokeh image. In each iteration,
the defocus clipping layer aims to prevent the scaled defocus values from exceeding the acceptable range of the subsequent network, and the thresholding layer produces a mask to replace the rendered result within clipping areas with the bilinear upsampled input bokeh image.
}
\label{fig:iunet}
\end{figure}

\vspace{3pt}  
\noindent\textbf{IUNet} iteratively upsamples the low-resolution bokeh image $B^{\,lr}_{nr}$ by a factor of $2$ until reaching the original resolution (Fig.~\ref{fig:iunet}). To avoid the fuzziness around in-focus areas caused by direct bilinear upsampling, we use the original high-resolution input as a guidance map. In each iteration, it is resized to twice the resolution of the input bokeh image. To match the increasing blur size during the iteration, we 
also need to dynamically adjust the values of the defocus map. Specifically, we once again use the adaptive resizing layer, and the downscale factor of each iteration $t$ is set as
\begin{equation}
    w^{(t)} = \frac{1}{2}\,w^{(t-1)},\quad t=1,...,T\,.
\end{equation}
However, with the progress of iteration, the scaled defocus values may exceed the acceptable range $[-\hat{R},\hat{R}]$ of the neural network. Fortunately, the fuzziness caused by direct bilinear upsampling is unnoticeable for the areas with large amount of bokeh blur. Thus, we can just refine the areas whose defocus values are in the range. To this end, we first clip the out-of-range defocus values to ensure that the subsequent network can work without collapse. Then, we threshold the dilated defocus map $S^d$ to produce a mask, which indicates the effectively rendered areas without defocus clipping. In these areas, we use the output of the network, while for the rest of the areas, we use the input bokeh image after bilinear upsampling. Here, we use $S^d$ instead of $S$ because the negative effects caused by defocus clipping will spread during the rendering.
The detailed calculation of $S^d$ is in the supplementary material. Overall, with the increase of iteration, the resolution of the bokeh image will be higher, but the effective area refined by the network will be smaller. In other words, areas nearby
the focal plane will be refined more times.

Finally, following alpha blending~\cite{lu2019indices,xu2017deep}, we use the predicted error map $E$ to fuse the bokeh results of the classical renderer $B_{cr}$ and that of the neural renderer $B_{nr}$:
\begin{equation}
    B = (1-E) \cdot B_{cr} + E \cdot B_{nr}\,.
\end{equation}

\begin{table*}
    \setlength{\abovecaptionskip}{5pt}
	\centering
	\caption{Quantitative results on the BLB dataset. Different levels correspond to different blur parameters of bokeh images, \eg, ``Level 1'' denotes that the blur parameter is $10$, and ``Level 5'' denotes that the blur parameter is $50$. The best performance is in \textbf{boldface}.} 
	\resizebox{1.0\linewidth}{!}{
	\renewcommand\arraystretch{1.0}
	\begin{tabular}{lccccccccccccccc}
		\toprule
		\multicolumn{1}{l}{\multirow{2}{*}[-0.5ex]{Methods}} & \multicolumn{3}{c}{Level 1} & \multicolumn{3}{c}{Level 2} & \multicolumn{3}{c}{Level 3} & \multicolumn{3}{c}{Level 4} & \multicolumn{3}{c}{Level 5} \\
		\cmidrule(r){2-4} \cmidrule(r){5-7} \cmidrule(r){8-10} \cmidrule(r){11-13} \cmidrule(r){14-16} 
		~ & PSNR & SSIM & Time(s) & PSNR & SSIM & Time(s) & PSNR & SSIM & Time(s) & PSNR & SSIM & Time(s) & PSNR & SSIM & Time(s) \\
		\midrule
		VDSLR~\cite{yang2016virtual} & 41.13 & 0.9891 & 0.06 & 39.15 & 0.9848 & 0.23 & 37.64 & 0.9812 & 0.53 & 36.48 & 0.9783 & 0.97 & 35.57 & 0.9760 & 1.55 \\
		SteReFo~\cite{busam2019sterefo} &
		37.21 & 0.9831 & 0.13 & 35.28 & 0.9818 & 0.60 & 33.99 & 0.9813 & 1.69 & 32.94 & 0.9809 & 3.74 & 32.12 & 0.9805 & 6.87 \\
		RVR~\cite{zhang2019synthetic} & 32.35 & 0.9648 & 0.10 & 32.00 & 0.9321 & 0.43 & 28.36 & 0.9011 & 1.11 & 25.80 & 0.8775 & 2.30 & 23.94 & 0.8596 & 4.12 \\
		RVR$^\dagger$~\cite{zhang2019synthetic} & 37.15 & 0.9836 & 0.13 & 38.55 & 0.9880 & 0.62 & 35.56 & 0.9854 & 1.82 & 33.03 & 0.9815 & 3.97 & 31.15 & 0.9774 & 7.21 \\
		\midrule
		DeepLens~\cite{wang2018deeplens} & 33.68 & 0.9679 & 0.14 & 31.43 & 0.9603 & 0.14 & 30.16 & 0.9564 & \textbf{0.14} & 29.30 & 0.9539 & 0.14 & 28.68 & 0.9521 & 0.14 \\
		DeepFocus~\cite{xiao2018deepfocus} & 38.92 & 0.9900 & 0.71 & 36.13 & 0.9857 & 0.71 & 31.47 & 0.9623 & 0.71 & 25.55 & 0.9089 & 0.71 & 21.04 & 0.8227 & 0.71 \\
		DeepFocus$^\dagger$~\cite{xiao2018deepfocus} & 38.92 & 0.9900 & 0.71 & 35.74 & 0.9861 & 0.49 & 34.21 & 0.9833 & 0.22 & 33.21 & 0.9809 & \textbf{0.13} & 32.44 & 0.9788 & \textbf{0.09} \\
		\midrule
		Ours (CR) & 41.32 & 0.9900 & \textbf{0.03} & 39.51 & 0.9877 & \textbf{0.10} & 38.35 & 0.9868 & 0.20 & 37.53 & 0.9864 & 0.34 & 36.86 & 0.9862 & 0.52 \\
		Ours (NR) & 40.41 & 0.9905 & 0.13 & 40.16 & 0.9904 & 0.13 & 39.21 & 0.9896 & \textbf{0.14} & 38.01 & 0.9884 & 0.16 & 37.20 & 0.9875 & 0.16 \\
		Ours &
		\textbf{43.30} & \textbf{0.9932} & 0.16 & \textbf{42.21} & \textbf{0.9924} & 0.23 & \textbf{41.02} & \textbf{0.9915} & 0.34 & \textbf{39.78} & \textbf{0.9906} & 0.50 & \textbf{38.80} & \textbf{0.9898} & 0.68 \\
		\bottomrule
		\vspace{-10mm}
	\end{tabular}
	}
	\label{tab:BLB}
\end{table*}

\vspace{3pt}
\noindent\textbf{Loss Functions.} We train ARNet and IUNet separately. When training ARNet, the adaptive resizing layer is unused. $B$ is fused by $B_{cr}$ and $B^{\,lr}_{nr}$. The loss is defined by
\begin{align}\label{eqn:loss_arnet}
    \mathcal{L}_{AR} & = \mathcal{L}_{\ell_1}(B,B^*) + \mathcal{L}_{\ell_1}(\nabla B,\nabla B^*) \nonumber\\
    & + \mathcal{L}_{\ell_1}(B^{\,lr}_{nr},B^*) + \mathcal{L}_{\ell_1}(\nabla B^{\,lr}_{nr},\nabla B^*) \nonumber\\
    & + \lambda_{bce}\,\mathcal{L}_{bce}(E,E^*)\,,
\end{align}
where ground-truth maps are marked with a superscript $*$. $\nabla$ denotes the image gradient. $\lambda_{bce}$ is empirically set to $0.1$. When training IUNet, we freeze ARNet and use the following loss:
\begin{align}
    \mathcal{L}_{IU} & = \mathcal{L}_{\ell_1}(B,B^*) + \mathcal{L}_{\ell_1}(\nabla B,\nabla B^*) \nonumber\\
    & + \mathcal{L}_{\ell_1}(B_{nr},B^*) + \mathcal{L}_{\ell_1}(\nabla B_{nr},\nabla B^*)\,.
\end{align}
Note that, for fast convergence, we supervise the training of both ARNet and IUNet with the intermediate result $B^{\,lr}_{nr}$ or $B_{nr}$, aside from the final result $B$.

\vspace{3pt}
\noindent\textbf{Implementations.} 
Our implementation is based on PyTorch~\cite{paszke2017automatic}.
To train the neural renderer, we synthesize a bokeh dataset using a simplified ray tracing method. 
This dataset contains $150$ scenes. For each scene, it consists of an all-in-focus image, a disparity map ranging from $0$ to $1$, and a stack of bokeh images with $2$ blur parameters ($12$, $24$), $20$ refocused disparities ($0.05$, $0.1$, ..., $1$), and $5$ gamma values ($1$, $2$, ..., $5$). 
We use the data with the blur parameter of $12$ for ARNet training and $24$ for IUNet training. 
We follow the same data pre-processing configurations as in~\cite{xiao2018deepfocus}. To improve the generalization, we additionally augment the input disparity map with random gaussian blur, dilation and erosion. The acceptable defocus ranges of ARNet and IUNet are both set to $[-12, 12]$ for training and $[-10, 10]$ for inference. Both networks are trained for $50$ epochs with a batch size of $16$. The learning rate is set to $10^{-4}$. Adam optimizer~\cite{kingma2015adam} is used for optimization. 
All experiments are conducted on an NVIDIA GeForce GTX 1080 Ti GPU.

\section{Experiments}

\subsection{Test Data}
For all test data, without loss of generality, we assume that the aperture shape is circular and the gamma value is $2.2$ to create a level playing field for different methods. The disparity maps of all datasets are normalized to $[0,1]$ range.

\vspace{3pt}
\noindent\textbf{BLB} contains $500$ test samples synthesized by Blender 2.93~\cite{blender}. Specifically, we download $10$ 3D scene models of Blender splash screens from different versions~\cite{gallery}. For each scene model, we use Cycles Engine~\cite{blender} to render an all-in-focus image, a disparity map, and a stack of bokeh images with $5$ blur parameters and $10$ refocused disparities.
The image resolution is set to $1920\times1080$. 

\vspace{3pt}
\noindent\textbf{EBB400} contains $400$ wide and shallow DoF image pairs which are randomly selected from EBB!~\cite{ignatov2020rendering}. 
For each sample, we predict a disparity map by MiDaS~\cite{ranftl2020towards}, and manually label a bounding box referred to the in-focus areas, so that we can obtain the refocused disparity by taking the median value of the disparity map within the bounding box~\cite{peng2021interactive}. The image resolution is about $1536\times1024$.

\vspace{3pt}
\noindent\textbf{IPB} contains $40$ images captured by iPhone 12 Portrait mode. For each scene, we first export an all-in-focus image and a bokeh image post-processed by the Portrait mode from iPhone 12. Then, using the online photo editor Photopea~\cite{photopea}, we can further extract a disparity map and an irradiance map from the bokeh image. All images are shot vertically with the resolution of $3024\times4032$.

\begin{figure*}
    \setlength{\abovecaptionskip}{3pt}
    \setlength{\belowcaptionskip}{-10pt}
    \small
	\centering
	\renewcommand\arraystretch{.5}
	\begin{tabular}{*{5}{c@{\hspace{.6mm}}}}
        \begin{tikzpicture}[spy using outlines={rectangle,red,magnification=5,width=0.094\linewidth,height=0.055\linewidth,every spy on node/.append style={thick}}]
        \node[inner sep=0]{\includegraphics[trim={50pt 0 250pt 50pt},clip,width=0.193\linewidth]{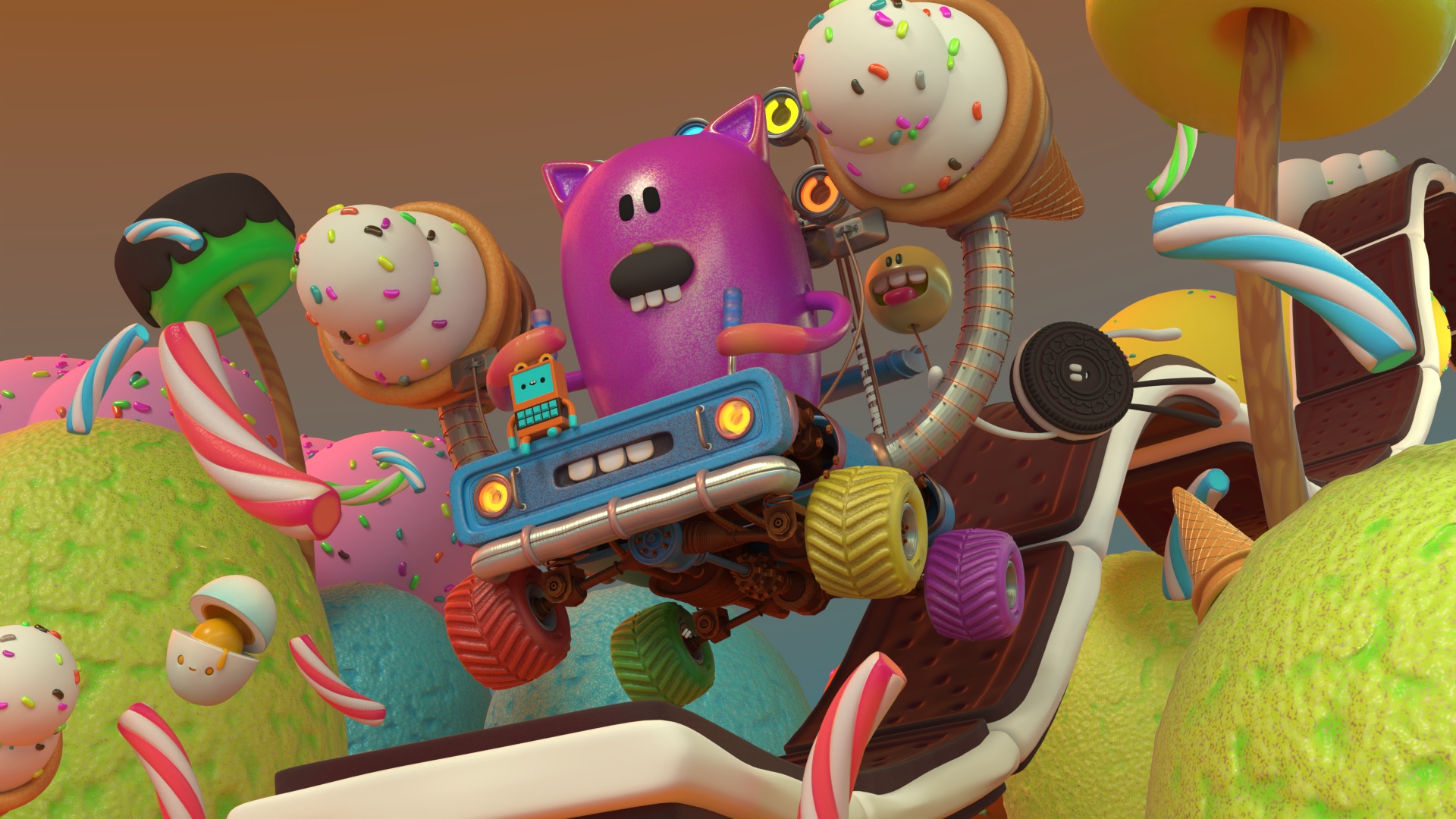}};
        \spy[red] on (0.53,0.36)in node at (-0.85,-1.59);
        \spy[green] on (0.53,-0.85) in node at (0.85,-1.59);
        \end{tikzpicture} &
        
        \begin{tikzpicture}[spy using outlines={rectangle,red,magnification=5,width=0.094\linewidth,height=0.055\linewidth,every spy on node/.append style={thick}}]
        \node[inner sep=0]{\includegraphics[trim={50pt 0 250pt 50pt},clip,width=0.193\linewidth]{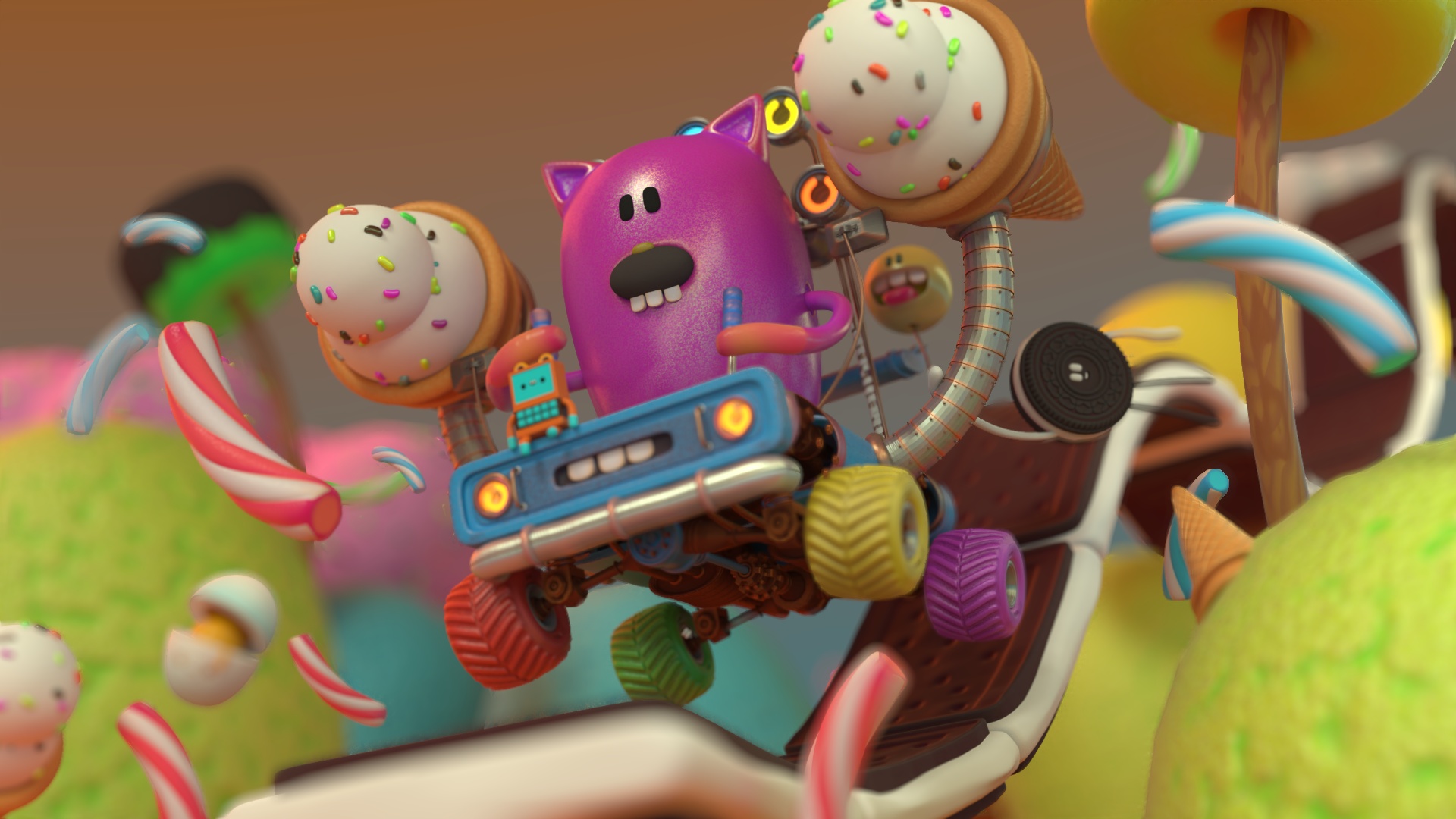}};
        \spy[red] on (0.53,0.36)in node at (-0.85,-1.59);
        \spy[green] on (0.53,-0.85) in node at (0.85,-1.59);
        \end{tikzpicture} &
        
        \begin{tikzpicture}[spy using outlines={rectangle,red,magnification=5,width=0.094\linewidth,height=0.055\linewidth,every spy on node/.append style={thick}}]
        \node[inner sep=0]{\includegraphics[trim={50pt 0 250pt 50pt},clip,width=0.193\linewidth]{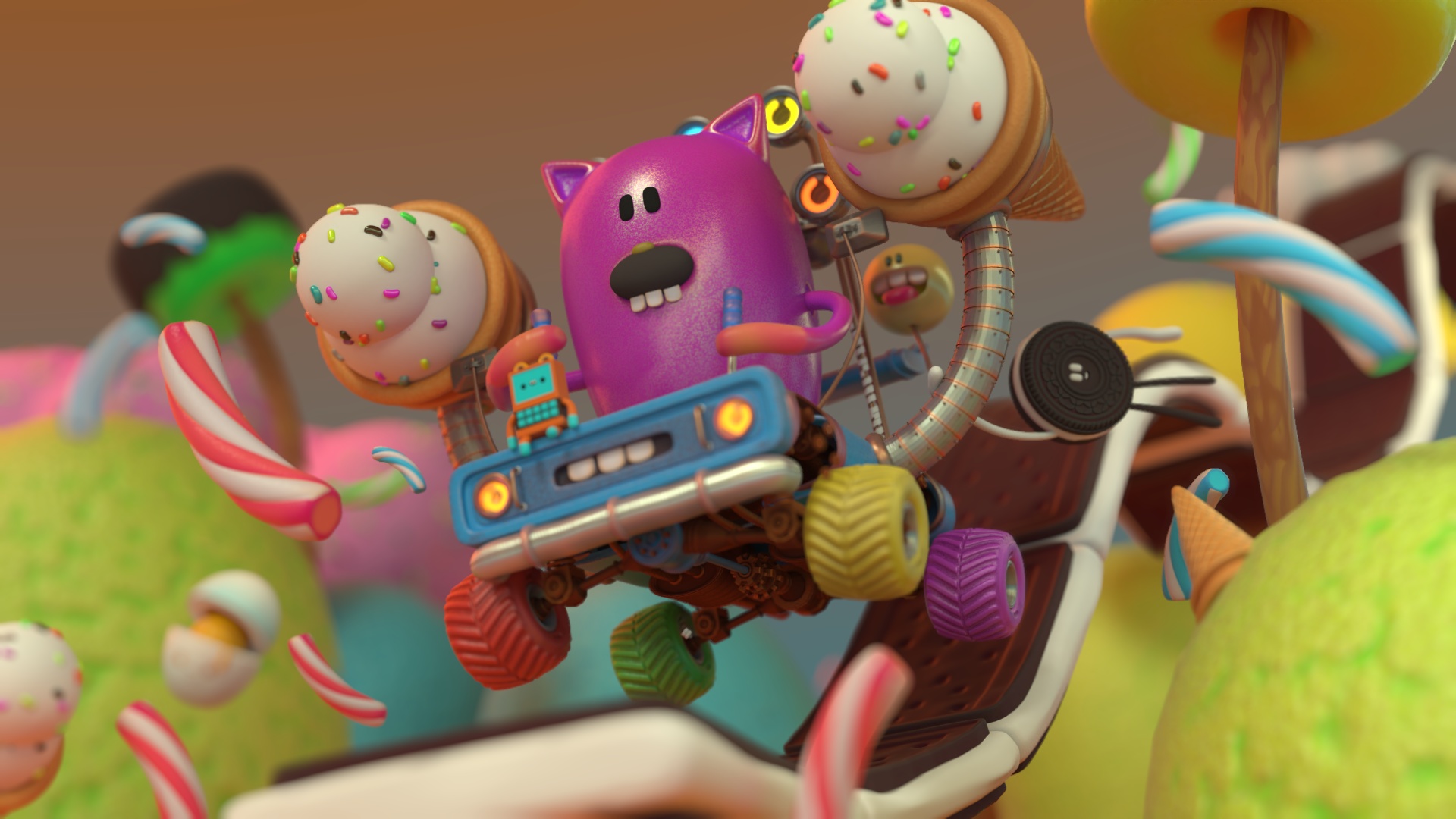}};
        \spy[red] on (0.53,0.36)in node at (-0.85,-1.59);
        \spy[green] on (0.53,-0.85) in node at (0.85,-1.59);
        \end{tikzpicture} &
        
        \begin{tikzpicture}[spy using outlines={rectangle,red,magnification=5,width=0.094\linewidth,height=0.055\linewidth,every spy on node/.append style={thick}}]
        \node[inner sep=0]{\includegraphics[trim={50pt 0 250pt 50pt},clip,width=0.193\linewidth]{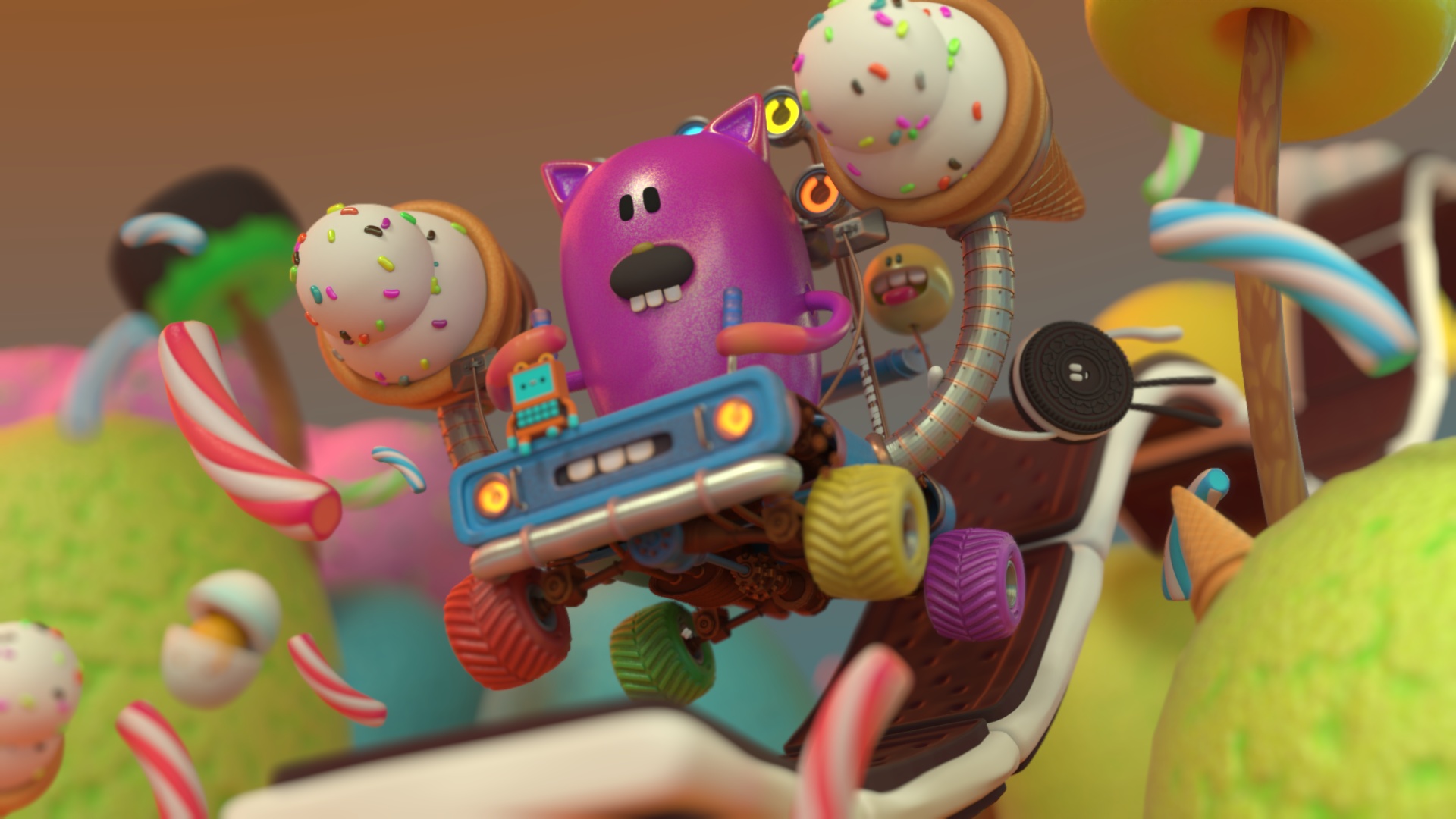}};
        \spy[red] on (0.53,0.36)in node at (-0.85,-1.59);
        \spy[green] on (0.53,-0.85) in node at (0.85,-1.59);
        \end{tikzpicture} &
        
        \begin{tikzpicture}[spy using outlines={rectangle,red,magnification=5,width=0.094\linewidth,height=0.055\linewidth,every spy on node/.append style={thick}}]
        \node[inner sep=0]{\includegraphics[trim={50pt 0 250pt 50pt},clip,width=0.193\linewidth]{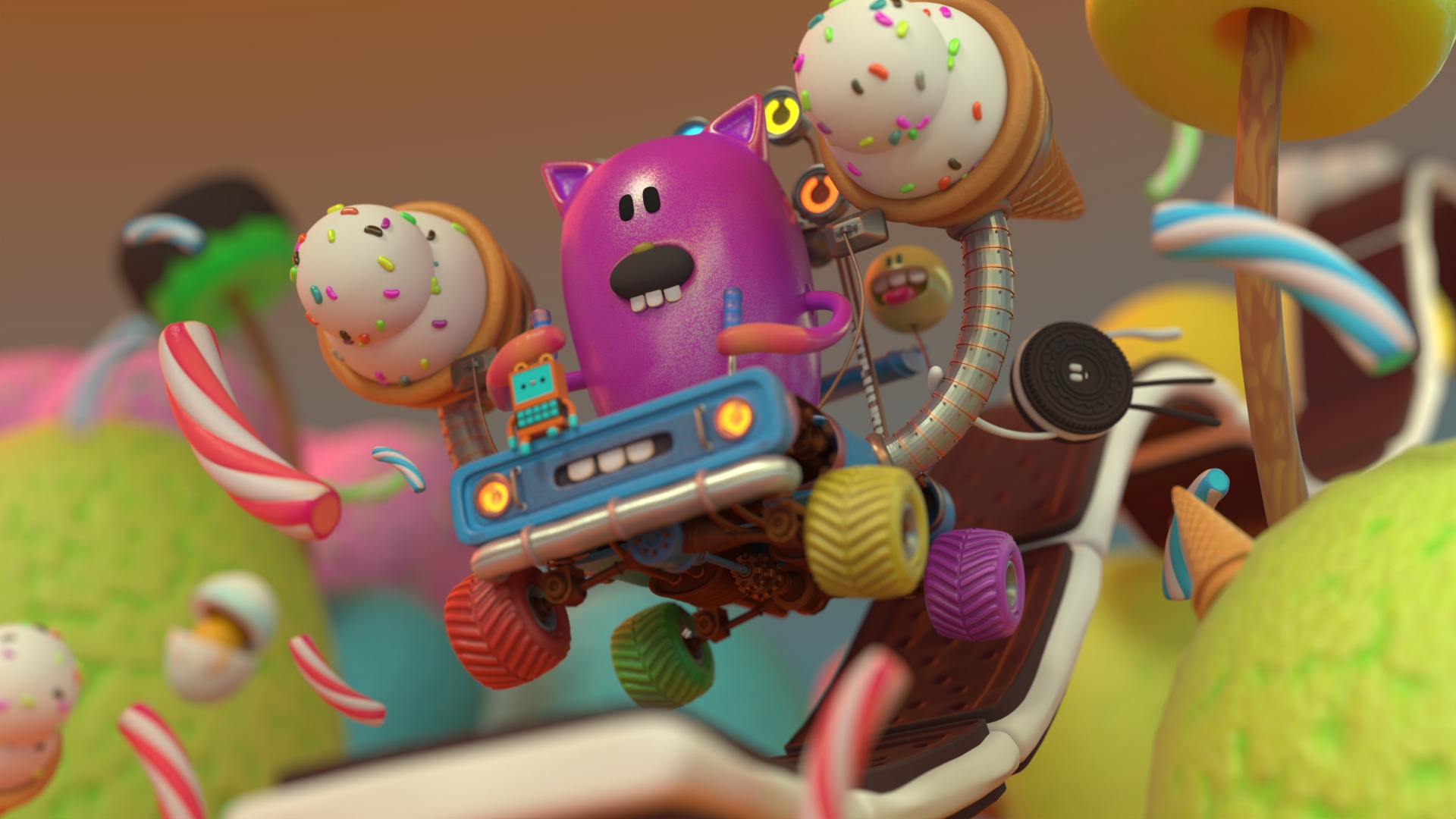}};
        \spy[red] on (0.53,0.36)in node at (-0.85,-1.59);
        \spy[green] on (0.53,-0.85) in node at (0.85,-1.59);
        \end{tikzpicture} \\
        
        \noalign{\vskip 0.4mm}
        
        All-in-Focus & VDSLR~\cite{yang2016virtual} & SteReFo~\cite{busam2019sterefo} & RVR$^\dagger$~\cite{zhang2019synthetic} & GT \\
        
        \noalign{\vskip 1mm}
        
        \begin{tikzpicture}[spy using outlines={rectangle,red,magnification=5,width=0.094\linewidth,height=0.055\linewidth,every spy on node/.append style={thick}}]
        \node[inner sep=0]{\includegraphics[trim={50pt 0 250pt 50pt},clip,width=0.193\linewidth]{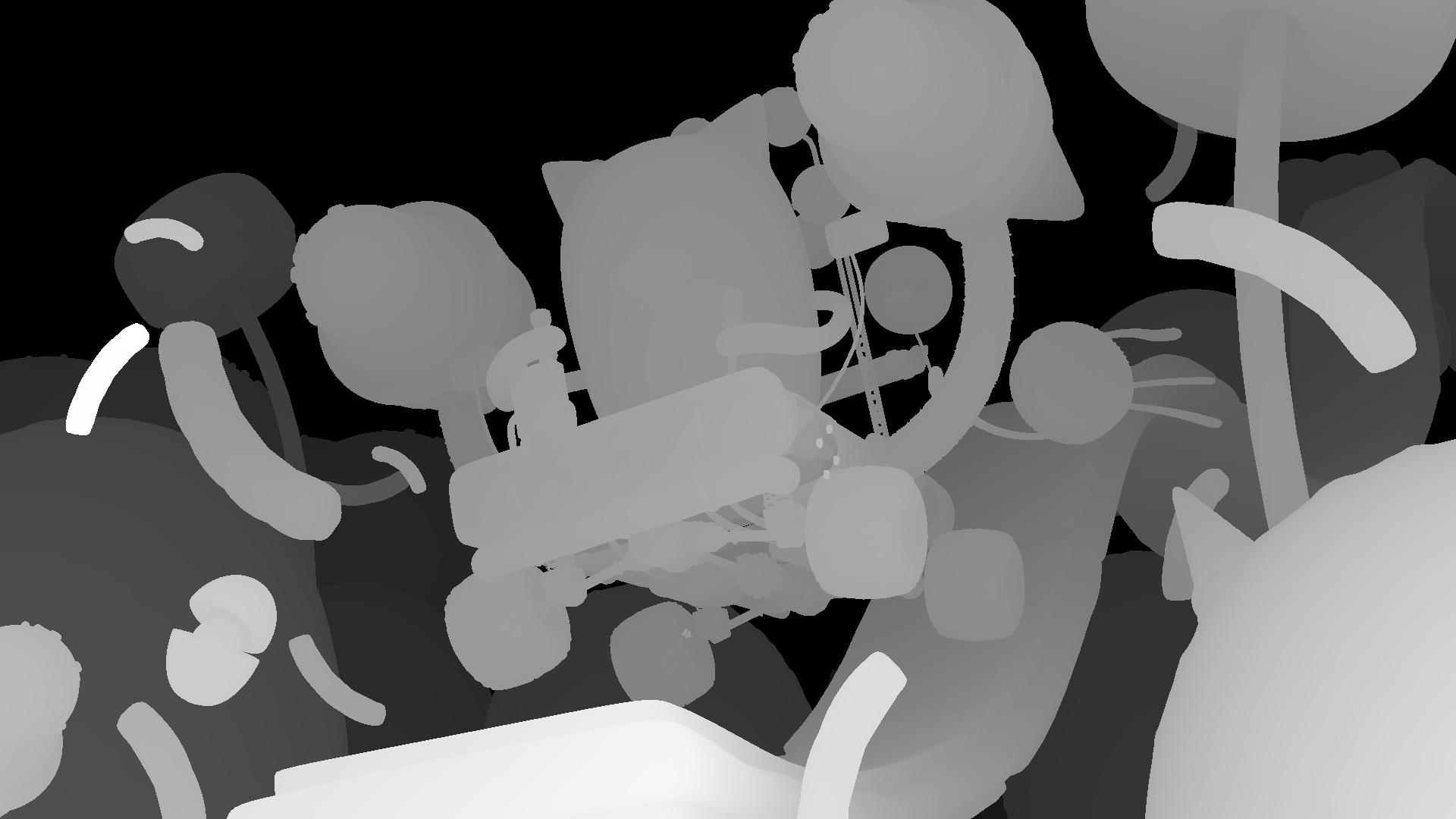}};
        \spy[red] on (0.53,0.36)in node at (-0.85,-1.59);
        \spy[green] on (0.53,-0.85) in node at (0.85,-1.59);
        \draw (0.05,0.5) pic[thick,yellow] {cross=2pt};
        \end{tikzpicture} &
        
        \begin{tikzpicture}[spy using outlines={rectangle,red,magnification=5,width=0.094\linewidth,height=0.055\linewidth,every spy on node/.append style={thick}}]
        \node[inner sep=0]{\includegraphics[trim={50pt 0 250pt 50pt},clip,width=0.193\linewidth]{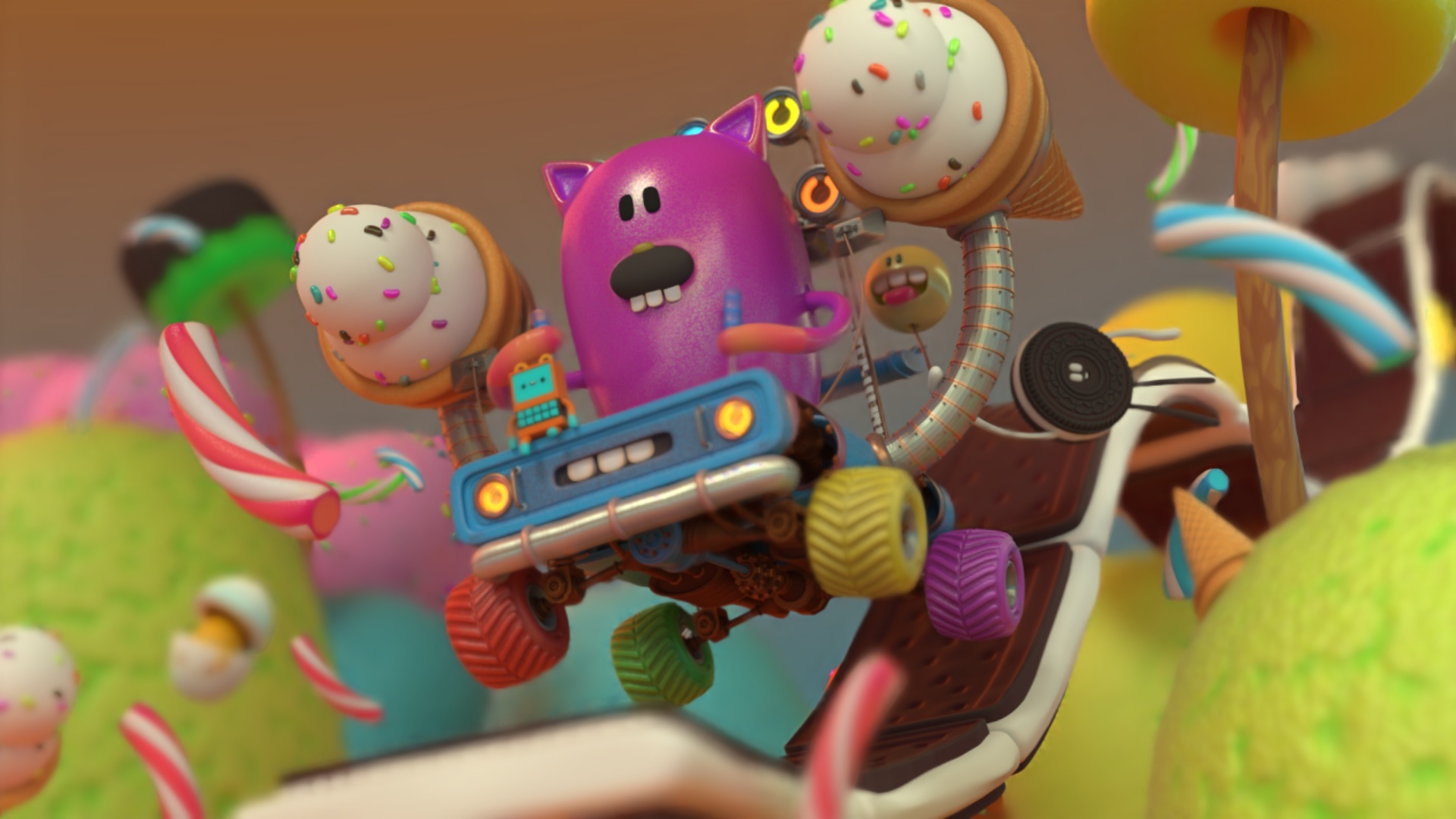}};
        \spy[red] on (0.53,0.36)in node at (-0.85,-1.59);
        \spy[green] on (0.53,-0.85) in node at (0.85,-1.59);
        \end{tikzpicture} &
        
        \begin{tikzpicture}[spy using outlines={rectangle,red,magnification=5,width=0.094\linewidth,height=0.055\linewidth,every spy on node/.append style={thick}}]
        \node[inner sep=0]{\includegraphics[trim={50pt 0 250pt 50pt},clip,width=0.193\linewidth]{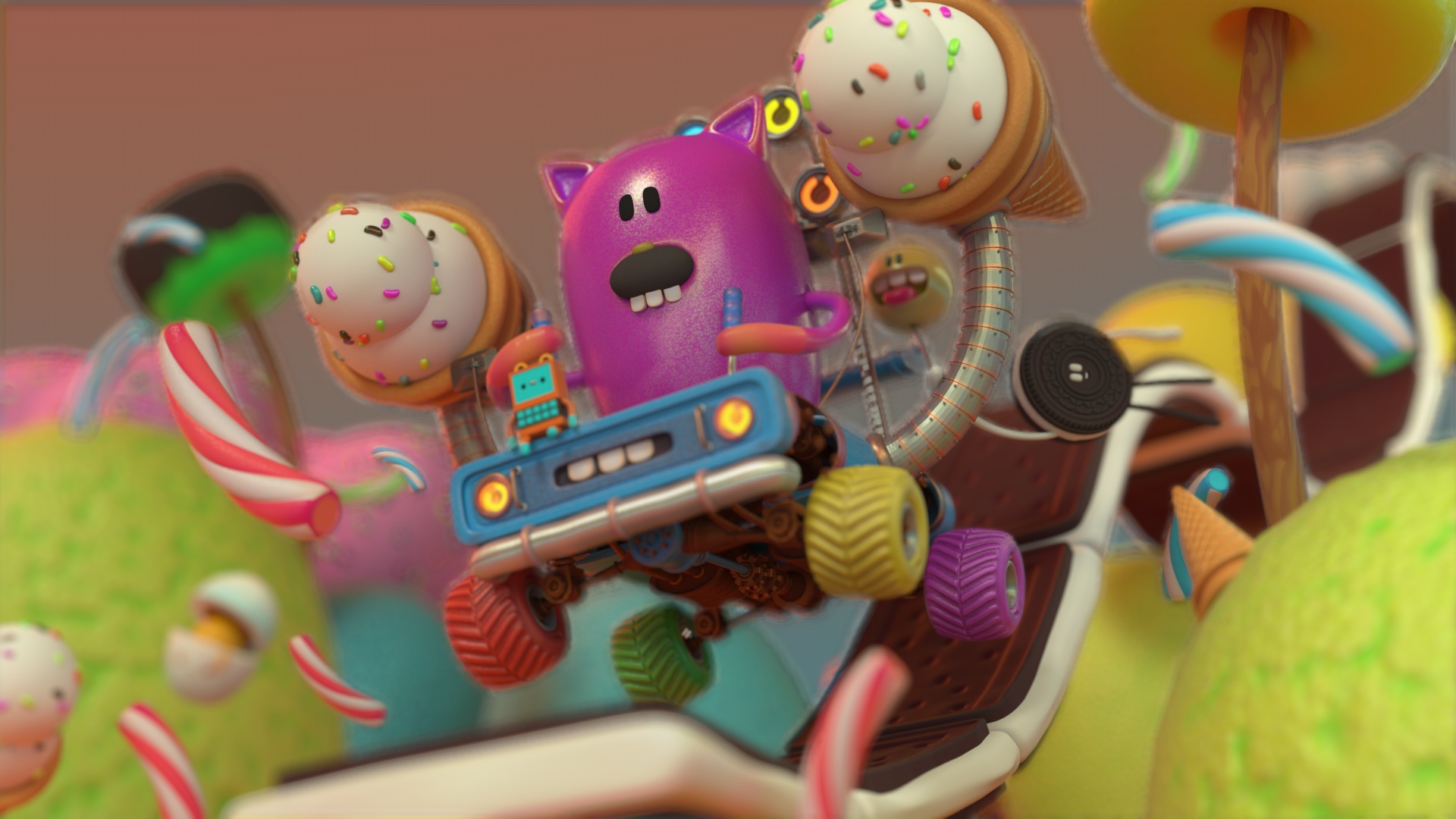}};
        \spy[red] on (0.53,0.36)in node at (-0.85,-1.59);
        \spy[green] on (0.53,-0.85) in node at (0.85,-1.59);
        \end{tikzpicture} &
        
        \begin{tikzpicture}[spy using outlines={rectangle,red,magnification=5,width=0.094\linewidth,height=0.055\linewidth,every spy on node/.append style={thick}}]
        \node[inner sep=0]{\includegraphics[trim={50pt 0 250pt 50pt},clip,width=0.193\linewidth]{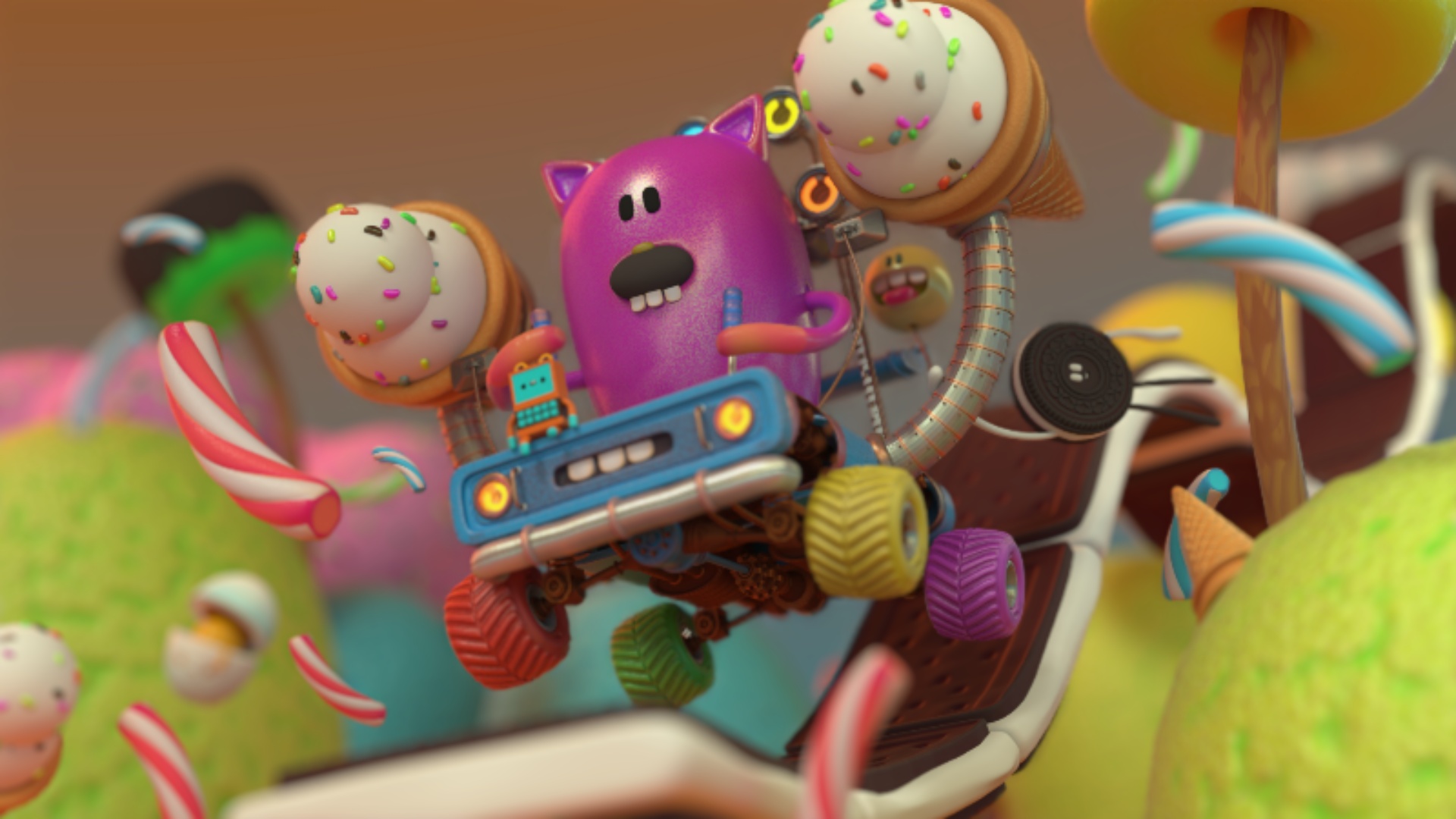}};
        \spy[red] on (0.53,0.36)in node at (-0.85,-1.59);
        \spy[green] on (0.53,-0.85) in node at (0.85,-1.59);
        \end{tikzpicture} &
        
        \begin{tikzpicture}[spy using outlines={rectangle,red,magnification=5,width=0.094\linewidth,height=0.055\linewidth,every spy on node/.append style={thick}}]
        \node[inner sep=0]{\includegraphics[trim={50pt 0 250pt 50pt},clip,width=0.193\linewidth]{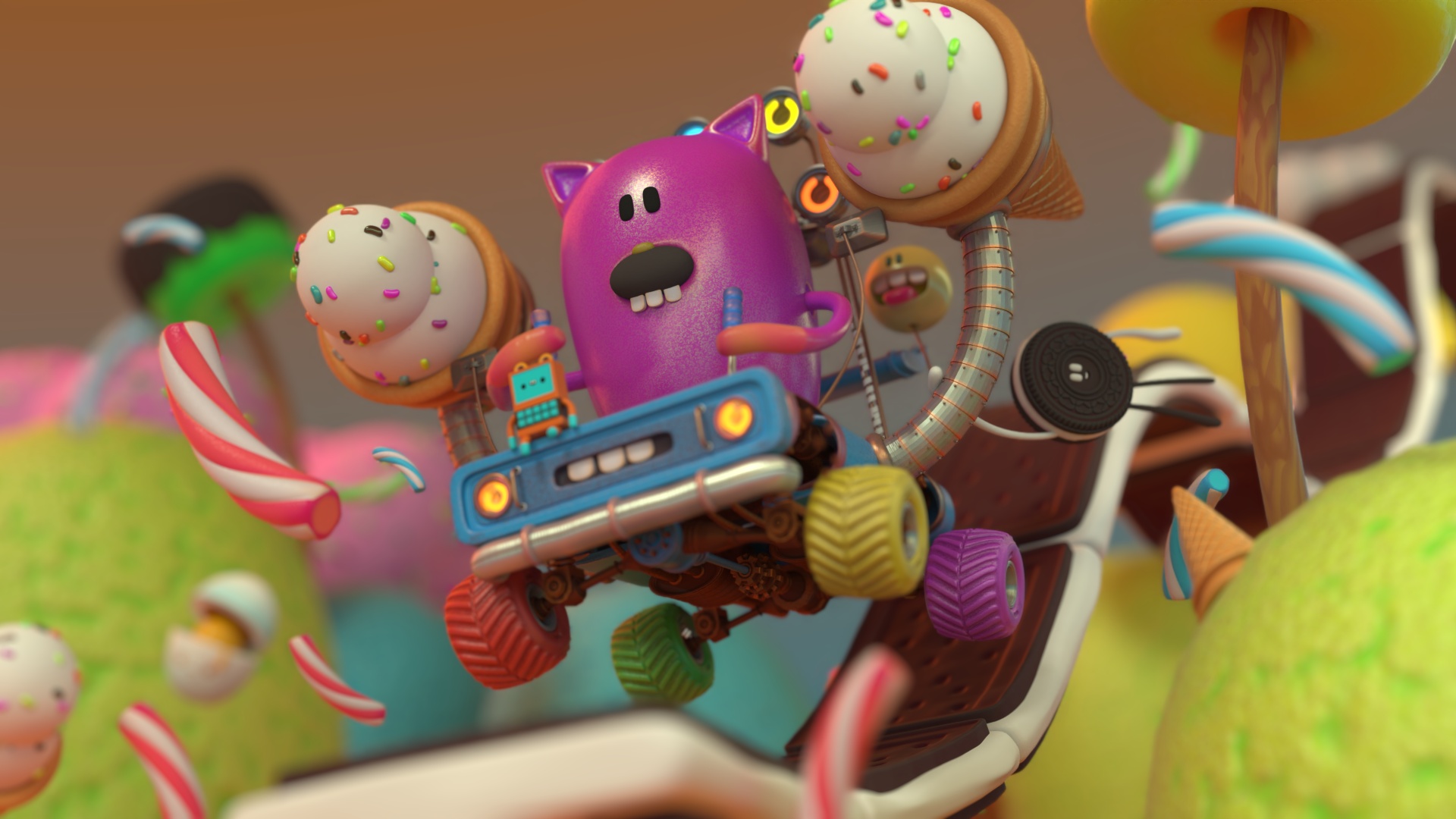}};
        \spy[red] on (0.53,0.36)in node at (-0.85,-1.59);
        \spy[green] on (0.53,-0.85) in node at (0.85,-1.59);
        \end{tikzpicture} \\
        
        \noalign{\vskip 0.4mm}
        
		Disparity & DeepLens~\cite{wang2018deeplens} & DeepFocus~\cite{xiao2018deepfocus} & DeepFocus$^\dagger$~\cite{xiao2018deepfocus} & Ours \\
	\end{tabular}
	\caption{Qualitative results on the BLB dataset. The rough refocused plane is labelled with a yellow cross on the disparity map.}
	\label{fig:BLB}
\end{figure*}

\subsection{Compared Methods}
We compare BokehMe with two types of methods: classical rendering methods and neural rendering methods. For simplicity, we represent them as ``C'' and ``N'' in the following. For a fair comparison, we provide the same disparity map for all methods, and we only preserve their bokeh rendering modules, while the others are discarded.

\vspace{3pt}
\noindent\textbf{VDSLR~\cite{yang2016virtual} (C)} is a pixel-wise pseudo ray tracing method accelerated by randomized intersection searching.

\vspace{3pt}
\noindent\textbf{SteReFo~\cite{busam2019sterefo} (C)} decomposes the image into layers according to the depth and renders the image from back to front.

\vspace{3pt}
\noindent\textbf{RVR~\cite{zhang2019synthetic} (C)} is similar to SteReFo. However, 
as discussed in~\cite{xian2021ranking}, 
original RVR lacks weight normalization, resulting in serious artifacts among different depth layers, so we add extra weight normalization as in SteReFo, and mark this modified method with superscript $\dagger$.

\vspace{3pt}
\noindent\textbf{DeepLens~\cite{wang2018deeplens} (N)} is trained on a homemade synthetic dataset and can generate high-resolution outputs.

\vspace{3pt}
\noindent\textbf{DeepFocus~\cite{xiao2018deepfocus} (N)} 
is trained on Unity data~\cite{unity}. As DeepFocus cannot handle large blur sizes, we apply the adaptive resizing layer proposed in our paper to the head of its model, and upsample its result to original resolution directly. Similarly, this modified method is marked with superscript $\dagger$.

\begin{figure}
\setlength{\abovecaptionskip}{-8pt}
\setlength{\belowcaptionskip}{-10pt}
\begin{center}
\includegraphics[width=\linewidth]{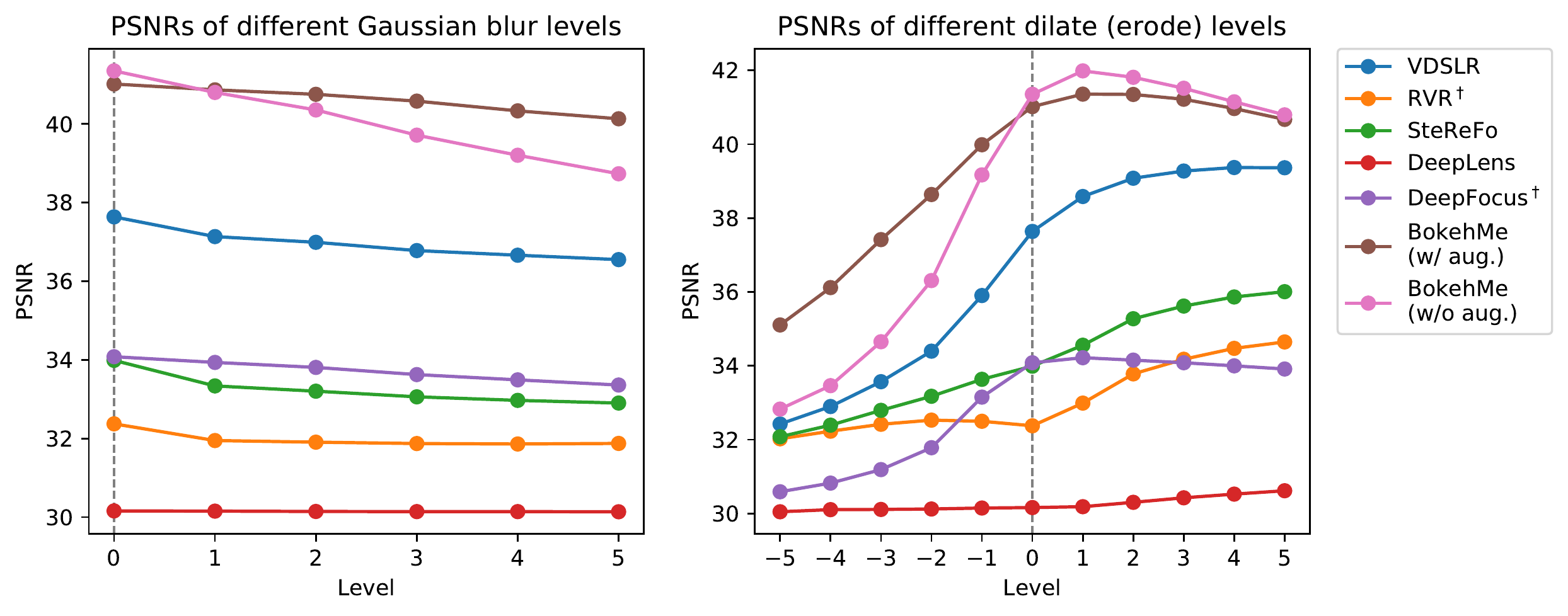}
\end{center}
\caption{
Evaluation on the BLB dataset with corrupting disparity maps. In the right chart, positive levels correspond to dilation levels while negative levels correspond to erosion levels.
}
\label{fig:corrupt1}
\end{figure}

\subsection{Zero-shot Cross-dataset Evaluation}
\label{sec:zero-shot}
Following~\cite{busam2019sterefo,xiao2018deepfocus,xu2018rendering}, we use PSNR and SSIM as metrics. We test BokehMe on the BLB dataset. As shown in Table~\ref{tab:BLB}, BokehMe achieves the best PSNR and SSIM scores compared with other state-of-the-art methods among all levels of blur, and our final model incorporating the classical renderer and the neural renderer outperforms either separate one, demonstrating strong complementarity between the two renderers. In addition, as the level of blur increases, the classical rendering methods become more time-consuming,
while the neural rendering methods maintain high efficiency.
We also show some visual results in Fig.~\ref{fig:BLB}. One can observe: (\rmnum{1}) 
The performance of classical methods degrades at depth discontinuities when the background is refocused on; (\rmnum{2}) DeepLens renders smooth results at depth discontinuities, but they seem not in line with the actual rendering; (\rmnum{3}) Compared with DeepFocus, DeepFocus$^\dagger$ avoids corruption in processing large blur sizes but generates blurry results around in-focus areas. (\rmnum{4}) Our approach renders most realistic bokeh effects for both in-focus and out-of-focus areas.

\begin{figure}
    \setlength{\abovecaptionskip}{3pt}
    \setlength{\belowcaptionskip}{-5pt}
    \small
	\centering
	\renewcommand\arraystretch{.5}
	\begin{tabular}{*{5}{c@{\hspace{.6mm}}}}
        \includegraphics[width=0.185\linewidth]{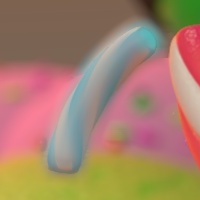} & \includegraphics[width=0.185\linewidth]{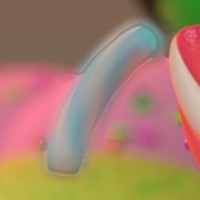} & 
        \includegraphics[width=0.185\linewidth]{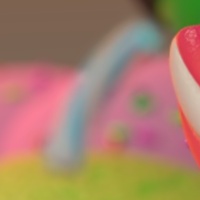} & 
        \includegraphics[width=0.185\linewidth]{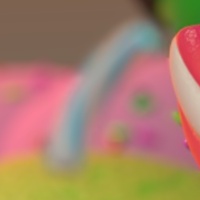} &
        \includegraphics[width=0.185\linewidth]{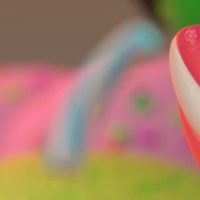} \\
        \noalign{\vskip 0.4mm}
        VDSLR & VDSLR$^{d_3}$ & Ours & Ours$^{d_3}$ & GT \\
    \end{tabular}
	\caption{Rendered results after dilating the disparity map on the BLB dataset. ``$d_3$'' means the level of dilation is $3$ (the kernel size is $7\times7$). The image originates from Fig.~\ref{fig:BLB}.
	}
	\label{fig:corrupt2}
\end{figure}

Since it is hard to acquire a disparity map in the real world, a common practice is to estimate one. Nevertheless, the predicted disparity map may be blurry and not align with the RGB image at  boundary. Therefore, we redo the ``Level 3'' experiment (Table.~\ref{tab:BLB}) by corrupting the disparity map with $5$ levels of gaussian blur, dilation and erosion, respectively. We also retrain BokehMe without disparity augmentation for extra comparison. As shown in Fig.~\ref{fig:corrupt1}, BokehMe trained with augmentation better adapts to imperfect disparity maps. Another interesting observation is that the moderate dilation improves the performance of most methods, especially for the classical ones. The reason may be that the dilated pixels that extend beyond the boundaries of the foreground object act as the occluded background pixels, leading to a significant improvement of metrics in case of background refocusing. However, as shown in Fig.~\ref{fig:corrupt2}, it causes more boundary artifacts at the same time.

\begin{table}
    \setlength{\abovecaptionskip}{5pt}
	\centering
	\caption{Quantitative results on the EBB400 dataset.} 
	\resizebox{1.0\linewidth}{!}{
	\renewcommand\arraystretch{1.0}
	\begin{tabular}{lcccccc}
		\toprule
		Methods & VDSLR & SteReFo & RVR$^\dagger$ & DeepLens & DeepFocus$^\dagger$ & Ours \\
		\midrule
		PSNR & 23.78 & 23.56 & 23.56 & 23.46 & 23.81 & \textbf{23.85} \\
		SSIM & 0.8738 & 0.8674 & 0.8690 & 0.8707 & 0.8754 & \textbf{0.8770} \\
		\bottomrule
		\vspace{-10mm}
	\end{tabular}
	}
	\label{tab:EBB400}
\end{table}

\begin{figure*}
    \setlength{\abovecaptionskip}{3pt}
    \setlength{\belowcaptionskip}{-5pt}
    \small
	\centering
	\renewcommand\arraystretch{.5}
	\begin{tabular}{*{6}{c@{\hspace{.6mm}}}}
        \begin{tikzpicture}\node[inner sep=0]{\includegraphics[width=0.16\linewidth]{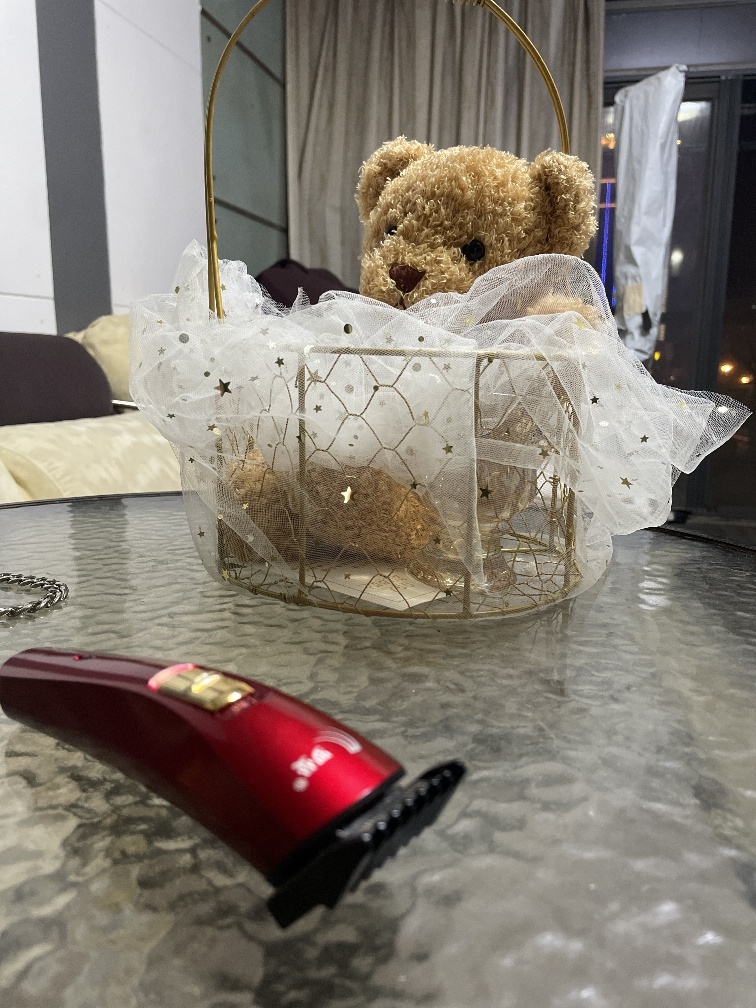}};
        \draw[color=red,thick] (-0.8,0.6) rectangle (-0.35,1.85);
        \draw[color=green,thick] (-1.385,-1.7) rectangle (0.45,-0.43);
        \end{tikzpicture} &
        \begin{tikzpicture}\node[inner sep=0]{\includegraphics[width=0.16\linewidth]{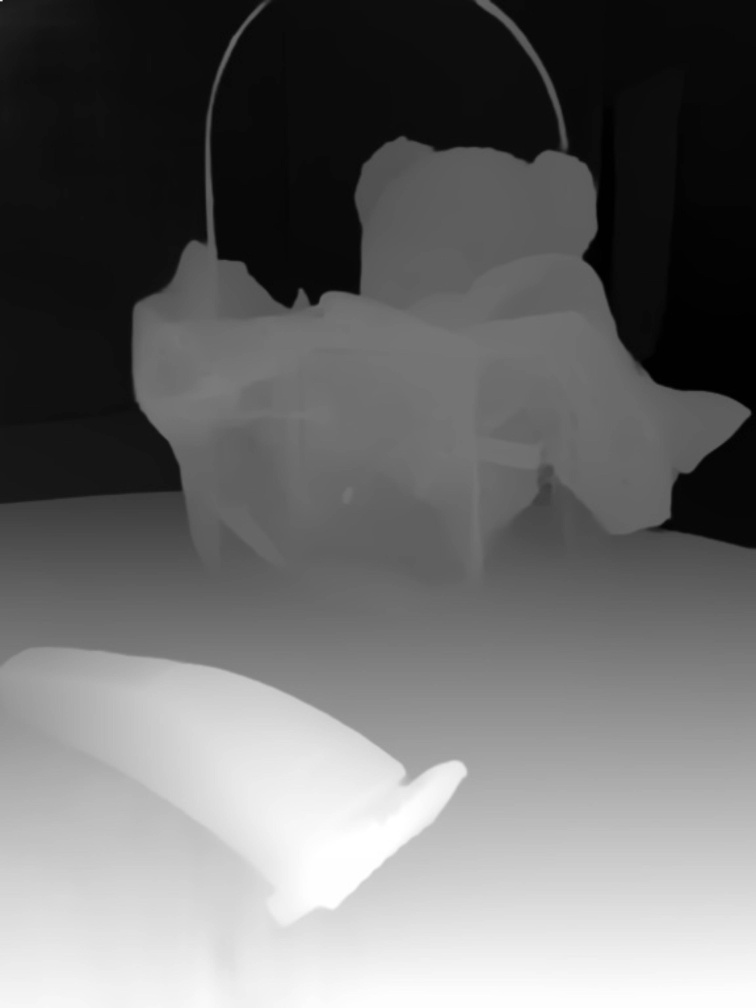}};
        \draw[color=red,thick] (-0.8,0.6) rectangle (-0.35,1.85);
        \draw[color=green,thick] (-1.385,-1.7) rectangle (0.45,-0.43);
        \draw (0.3,1.05) pic[thick,yellow] {cross=2pt};
        \end{tikzpicture} &
        \begin{tikzpicture}\node[inner sep=0]{\includegraphics[width=0.16\linewidth]{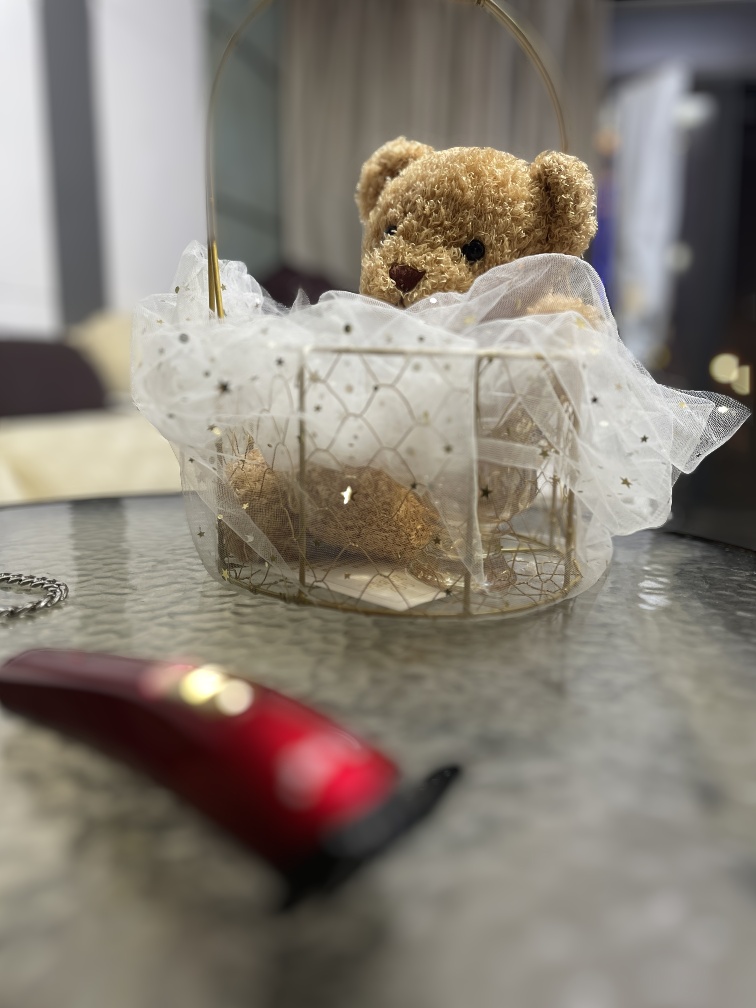}};
        \draw[color=red,thick] (-0.8,0.6) rectangle (-0.35,1.85);
        \draw[color=green,thick] (-1.385,-1.7) rectangle (0.45,-0.43);
        \end{tikzpicture} &
        \begin{tikzpicture}\node[inner sep=0]{\includegraphics[width=0.16\linewidth]{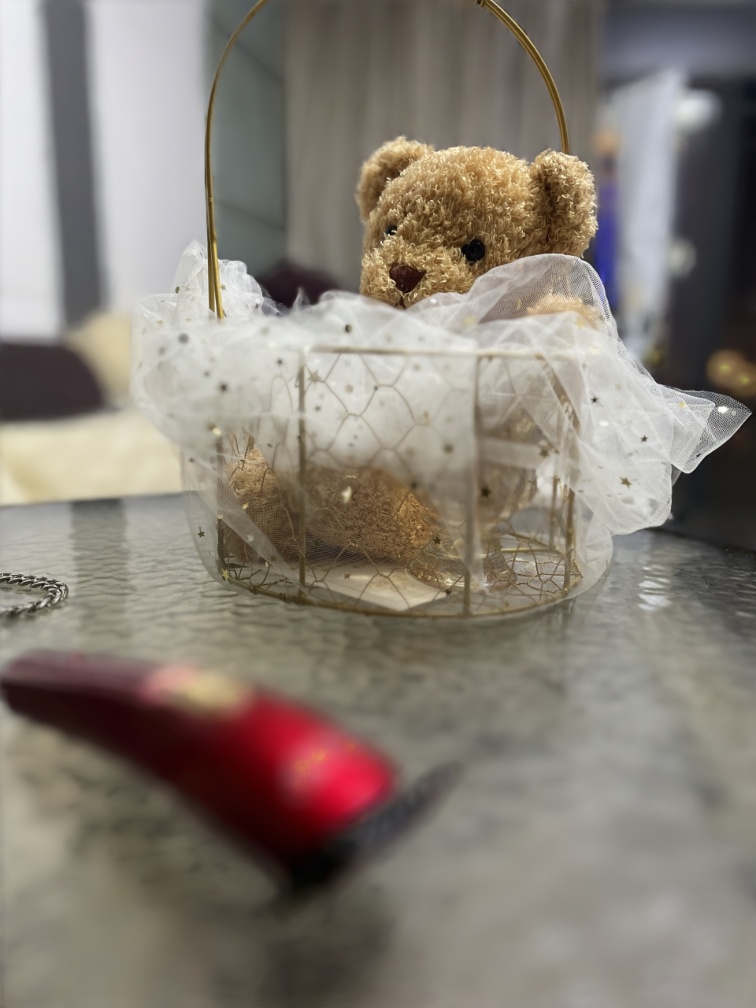}};
        \draw[color=red,thick] (-0.8,0.6) rectangle (-0.35,1.85);
        \draw[color=green,thick] (-1.385,-1.7) rectangle (0.45,-0.43);
        \end{tikzpicture} &
        \begin{tikzpicture}\node[inner sep=0]{\includegraphics[width=0.16\linewidth]{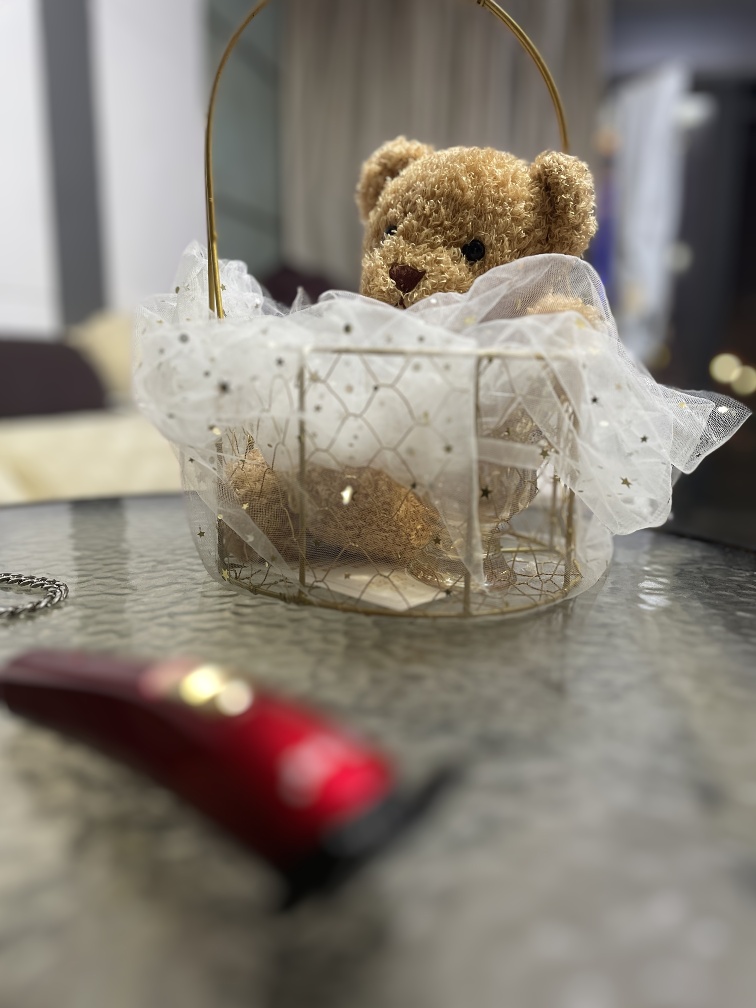}};
        \draw[color=red,thick] (-0.8,0.6) rectangle (-0.35,1.85);
        \draw[color=green,thick] (-1.385,-1.7) rectangle (0.45,-0.43);
        \end{tikzpicture} &
        \begin{tikzpicture}\node[inner sep=0]{\includegraphics[width=0.16\linewidth]{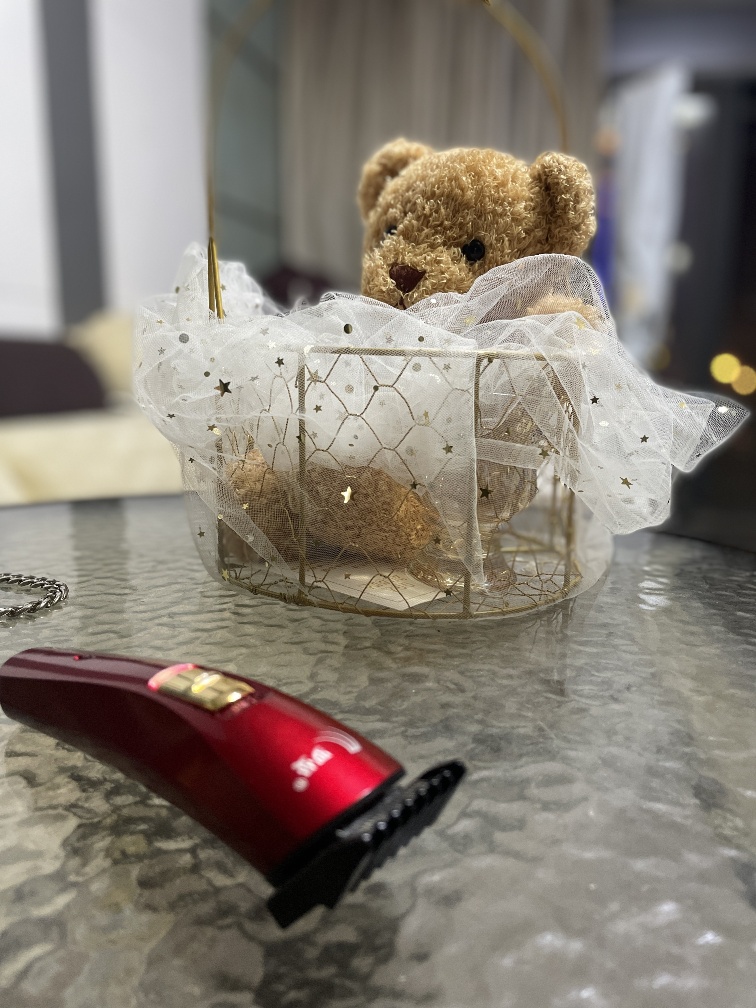}};
        \draw[color=red,thick] (-0.8,0.6) rectangle (-0.35,1.85);
        \draw[color=green,thick] (-1.385,-1.7) rectangle (0.45,-0.43);
        \end{tikzpicture} \\
        \noalign{\vskip 0.8mm}
        All-in-Focus & Disparity & VDSLR~\cite{yang2016virtual} & DeepLens~\cite{wang2018deeplens} & Ours & iPhone 12 Portrait \\
	\end{tabular}
	\caption{An example of user study on the IPB dataset. The rough refocused plane is labelled with a yellow cross on the disparity map.}
	\label{fig:user_study}
\end{figure*}
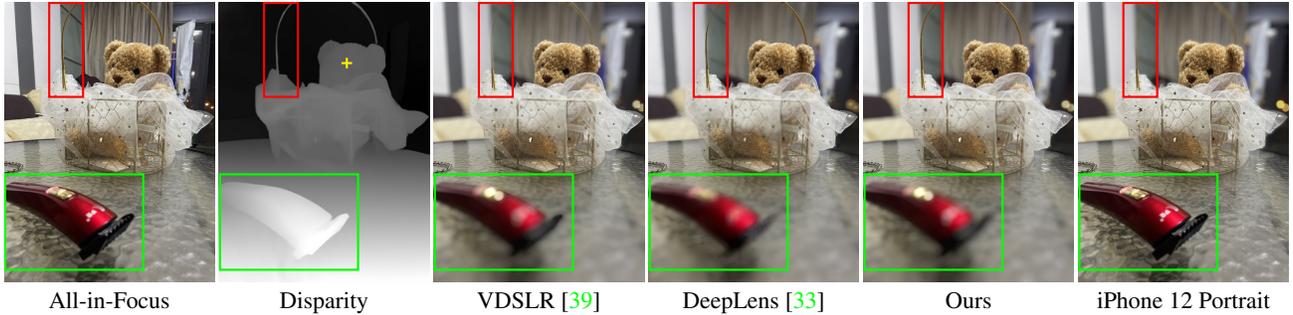

To further evaluate the generalization of the model given an imperfect disparity map as input, we compare different methods on the EBB400 dataset where disparity maps are predicted by MiDaS~\cite{ranftl2020towards}. As the blur parameter of each sample is unknown, we pick out the optimal value from $1$ to $100$ for each method. Despite the fact that color inconsistency and scene misalignment exist between the wide and shallow DoF image pairs, BokehMe still ranks the first in both metrics as shown in Table~\ref{tab:EBB400}. Refer to the supplementary material for qualitative results.

\subsection{Ablation Study}


\begin{table}
    \setlength{\abovecaptionskip}{5pt}
	\centering
	\caption{Ablation study of IUNet. ``B-Up'': bilinear upsampling; ``I-Up'': iterative upsampling by IUNet; ``Clip'': clipping of the signed defocus map; ``S-Fuse'': fusion with the mask thresholded by the signed defocus map; ``D-Fuse'': fusion with the mask thresholded by the dilated defocus map.}
	\resizebox{1.0\linewidth}{!}{
	\renewcommand\arraystretch{1.0}
	\begin{tabular}{l|ccccc|cc}
		\toprule
		No. & B-Up & I-Up & Clip & S-Fuse & D-Fuse & PSNR & SSIM \\
		\hline
		B1 & \checkmark & ~ & ~ & ~ & ~ & 37.30 & 0.9830 \\
		B2 & ~ & \checkmark & ~ & ~ & ~ & 23.23 & 0.8719 \\
		B3 & ~ & \checkmark & \checkmark & ~ & ~ & 38.55 & 0.9883 \\
		B4 & ~ & \checkmark & \checkmark & \checkmark & ~ & 39.19 & 0.9894 \\
		B5 & ~ & \checkmark & \checkmark & ~ & \checkmark & \textbf{39.21} & \textbf{0.9896} \\
		\bottomrule
	\end{tabular}
	}
	\vspace{-3mm}
	\label{tab:ablation2}
\end{table}


IUNet supports arbitrary-scale upsampling without losing quality.
To better understand how this outstanding characteristic is obtained, we conduct an ablation study on the BLB dataset (Level 3). Note that we only evaluate the neural renderer. Table~\ref{tab:ablation2} shows: (\rmnum{1}) Upsampling by IUNet without ``clipping'' will destroy the results because of out-of-range defocus values (B1 vs.\ B2 and B2 vs.\ B3); (\rmnum{2}) Using the low-resolution input bokeh image to compensate the clipping areas improves PSNR by $0.64$ dB (B3 vs.\ B4); (\rmnum{3}): Replacing the signed defocus map with dilated defocus map further improves metrics slightly (B4 vs.\ B5). Besides, we show in the supplementary material that this operation will provide a more natural boundary transition when the focal plane targets background.


\subsection{User Study}
Since PSNR and SSIM cannot fully reflect the actual quality of the rendered bokeh images, we conduct a user study on the IPB dataset. For all methods, the blur parameter and the refocused disparity are manually adjusted to match iPhone 12 Portrait mode. 
The study involves $53$ participants. From Table~\ref{tab:user_study} and Fig.~\ref{fig:user_study}, one can see that our approach is most favored with a clear boundary of in-focus objects and natural bokeh effects for foreground blur. Note that iPhone 12 Portrait mode can only produce bokeh effects for objects behind the focal point.

\begin{table}
    \setlength{\abovecaptionskip}{5pt}
	\centering
	\caption{User study results. Given a scene, participants are required to select one option from ``Good'', ``Normal'', and ``Bad'' for each anonymous method.} 
	\resizebox{0.85\linewidth}{!}{
	\renewcommand\arraystretch{1.0}
	\begin{tabular}{lcccc}
		\toprule
		Methods & iPhone 12 & VDSLR & DeepLens & Ours \\
		\midrule
		Good (\%) & 19.3 & 26.6 & 26.3 & \textbf{55.0} \\
		Normal (\%) & 29.3 & 47.7 & 45.0 & 38.5 \\
		Bad (\%) & 51.4 & 25.7 & 28.7 & \textbf{6.5} \\
		\bottomrule
		\vspace{-8mm}
	\end{tabular}
	}
	\label{tab:user_study}
\end{table}

\section{Discussion and Conclusion}
Classical rendering methods are flexible but suffer from artifacts at depth discontinuities. Neural rendering methods are capable of handling boundary artifacts but lack controllability and have difficulty in generating stunning bokeh balls in out-of-focus areas. To exploit the advantages of two paradigms, we propose BokehMe, a general framework that combines a classical renderer and a neural renderer. Extensive experiments illustrate that BokehMe can produce photo-realistic and highly controllable bokeh effects from an all-in-focus image and a potentially imperfect disparity map, demonstrating strong complementarity of classical rendering and neural rendering.

For BokehMe, the bokeh style can be controlled by changing the kernel shape of the classical renderer. It works for most scenes, however, if highlights happen to lie at the boundary of the error map, the bokeh style inconsistency may be noticeable. In addition, given a 8-bit digital image where bright lights exist in the scene, the gamma correction is insufficient to create prominent bokeh balls in out-of-focus areas. An ideal way is to transform the LDR image to an HDR image~\cite{debevec2008recovering}, by inverse tone mapping~\cite{cao2021brightness,kinoshita2019itm}, which is beyond the scope of this paper. Although similar effects can be achieved by forcibly enhancing the RGB values of input images, there is still room for improvement. We leave this in our future work.

\vspace{8pt}
\noindent\textbf{Acknowledgements.} This work was funded by Adobe.

\newpage

{\small

\bibliographystyle{ieee_fullname}
}

\end{document}